\documentclass[10pt,journal,compsoc]{IEEEtran}
\usepackage{xcolor,soul,framed} %,caption

\colorlet{shadecolor}{yellow}
\usepackage[pdftex]{graphicx}
\graphicspath{{../pdf/}{../jpeg/}}
\DeclareGraphicsExtensions{.pdf,.jpeg,.png}

\usepackage[cmex10]{amsmath}
\usepackage{array}
\usepackage{mdwmath}
\usepackage{mdwtab}
\usepackage{eqparbox}
\usepackage{url}
\usepackage{amsmath}
\usepackage{amsxtra}
\usepackage{amstext}
\usepackage{amssymb}
\usepackage{latexsym}
\usepackage{dsfont}
\usepackage{breqn}
\usepackage{graphicx}
\usepackage{enumerate}
\usepackage[ruled,vlined,linesnumbered,lined,boxed,commentsnumbered]{algorithm2e}
\usepackage{textcomp}
\graphicspath{ {./photo/} }
\newcommand{\ignore}[1]{}
\begin{document}
%\bstctlcite{IEEEexample:BSTcontrol}
\title{Efficient Implementation of a Multi-Layer Gradient-Free Online-Trainable Spiking Neural Network on FPGA}
\author{
{Ali~Mehrabi\IEEEauthorrefmark{1},~\IEEEmembership{Senior Member,~IEEE,}} {Yeshwanth~Bethi\IEEEauthorrefmark{1},~\IEEEmembership{Member,~IEEE,}} 
      {Andr\'e~van~Schaik\IEEEauthorrefmark{1},~\IEEEmembership{Fellow,~IEEE,}} {Andrew~Wabnitz\IEEEauthorrefmark{2}}, and {Saeed~Afshar\IEEEauthorrefmark{1},~\IEEEmembership{Member,~IEEE}}\\
{\{a.mehrabi, y.bethi, a.vanschaik, s.afshar\}@westernsydney.edu.au, andrew.wabnitz1@defence.gov.au}
\thanks{\IEEEauthorrefmark{1}International Center for Neuromorphic Systems (ICNS),\\The MARCS Institute of Brain, Behavior and Development,\\Western Sydney University.\\
\IEEEauthorrefmark{2}Defence Science and Technology Group, Department of Defence\\}

}
%\markboth{IEEE TRANSACTIONS 
%}{Roberg \MakeLowercase{\textit{et al.}}: High-Efficiency Diode and %Transistor Rectifiers}
%%%%%%%%%%%%%%%%%%%%%%%%%%%%%%%%%%%%%%%%%%%
\maketitle

% === ABSTRACT ====================================================================
% =================================================================================
\begin{abstract}
%\boldmath
This paper presents an efficient hardware implementation of the recently proposed Optimized Deep Event-driven Spiking Neural Network Architecture (ODESA). ODESA is the first network to have end-to-end multi-layer online local supervised training without using gradients and has the combined adaptation of weights and thresholds in an efficient hierarchical structure. This research shows that the network architecture and the online training of weights and thresholds can be implemented efficiently on a large scale in hardware. The implementation consists of a multi-layer Spiking Neural Network (SNN) and individual training modules for each layer that enable online self-learning without using back-propagation. By using simple local adaptive selection thresholds, a Winner-Takes-All (WTA) constraint on each layer, and a modified weight update rule that is more amenable to hardware, the trainer module allocates neuronal resources optimally at each layer without having to pass high-precision error measurements across layers.
All elements in the system, including the training module, interact using event-based binary spikes. The hardware-optimized implementation is shown to preserve the performance of the original algorithm across multiple spatial-temporal classification problems with significantly reduced hardware requirements.
\end{abstract}

% === KEYWORDS ====================================================================
% =================================================================================

% \fi
%\IEEEpeerreviewmaketitle

% ====================================================================
% ====================================================================
% ====================================================================

% === I. INTRODUCTION =============================================================
% =================================================================================
\section{Introduction}
\IEEEPARstart{A}{rtificial} Neural Networks (ANNs) and multi-layer perceptrons were developed as highly simplified models of biological neural computation through the use of distributed interconnected computing nodes, or neurons, which operate as a network, in contrast to the sequential architecture of conventional modern processors \cite{TURING, VONNEUMANN}. Deep ANNs have been developed, widely used, and optimized in the past two decades, resulting in significant advances in many scientific fields. As universal function approximators, ANNs can be applied to complex problems such as pattern recognition, classification, time series analysis, and speech recognition using the backpropagation algorithm \cite{HINTON} for training. 

During the same period, there has been a significant investigation and exploration of artificial Spiking Neural Networks (SNN), which are better models of biological neural networks by incorporating the spiking behavior of neurons observed in larger biological neural networks \cite{MAASS}. The investigation of SNNs is often motivated by the idea that the spiking behavior of biological nervous systems is functionally essential and provides computational and efficiency benefits \cite{LEVY, LAUGHLIN, VANRULLEN, ZADOR}. 

In contrast to ANNs, neurons in SNNs use precisely timed binary-valued pulse streams or spikes to transfer information. SNNs can perform sparse computations due to the inherent sparsity in their data. The ability to operate in an event-driven fashion, rather than the traditional synchronous clock-driven computational approach in ANNs, makes SNNs suitable for processing continuous-time spatio-temporal data. However, training SNNs is still an open research question, and a universal training algorithm akin to error backpropagation for ANNs is yet to be found.  

The spiking outputs generated by spiking neurons can be modeled as a train of Dirac's delta functions which do not have a derivative. The hard thresholding operation that is one of the key elements of function in spiking neuron models is also not differentiable. This non-differentiability of computations poses a fundamental challenge in assigning credit to earlier nodes in a network of spiking neurons to optimize synaptic weights. SpikeProp \cite{BOHTE}, Tempotron \cite{GUETIG}, Chronotron \cite{FLORIAN}, ReSuMe \cite{PONULAK}, and DL-ReSuMe \cite{TAHERKHANI} are some early methods introduced to apply gradient descent to train single-layered SNN models using various loss functions.  More recent works focused on approximating error backpropagation to SNN architectures, like using surrogate gradients for the different non-differentiable computations in an SNN  \cite{NEFTCI, ZENKE, BELLEC, BELLEC2, EVENTPROP}.  All the existing approximations of error backpropagation in SNNs batch data to accumulate gradients. They also require a symmetric backward data pathway to transfer continuous-valued gradients from the output layer through to the input layer to update neuronal weights in hidden layers. Some even rely on non-causal operations like Back Prop Through Time(BPTT) to update the synaptic weights. However, such non-local and non-causal operations in learning are not biologically plausible, and no such evidence of symmetric backward pathways in biological nervous systems is likely to be found. Despite the lack of bio-plausibility, error backpropagation methods have become popular tools to train SNNs for specific tasks. The error backpropagation methods are often computationally expensive, requiring energy-intensive GPUs to train them offline. 

Feedback alignment \cite{LILLICRAP} is one of the few alternatives to error backpropagation for SNNs, and it also requires passing continuous valued error values to each neuron. Local learning rules that do not require access to the weights of other neurons and communication of continuous-valued error gradients have been desirable for training SNNs. Variations of Spike Time-Dependent Plasticity (STDP) rules were applied to perform unsupervised feature extraction to classify spatio-temporal patterns \cite{DIEHL, TAVANEI, MASQUELIER, KHERADPISHEH, VIGNERON}. Mozafari et al. \cite{MOZAFARI} used reward-modulated STPD to perform object recognition. Paredes-Vall{\'e}s et al. \cite{PAREDES} used STDP rules to perform the optical-flow estimation. Local learning rules close to STDP, like Supervised Hebbian Learning \cite{LEGENSTEIN} and ReSuMe \cite{PONULAK2}, were also developed to perform supervised learning. However, multilayer versions \cite{SPOREA, TAHERKHANI2} rely on backpropagating continuous-valued feedback across hidden layers. 

In addition to training concerns, von Neumann computer architectures are not well suited for SNN implementations due to the massive parallelism inherent in an SNN network, where a large number of neurons must be processed simultaneously. While graphics processing units (GPUs) can implement parallelism to some extent, the kernel-launch programming paradigm makes them unsuitable for these applications. On the other hand, Field Programmable Gate Arrays (FPGAs) provide flexibility in designing parallel processing and re-configurable hardware architectures.

In many applications, SNNs can provide significant efficiency in power consumption due to the sparsity of inter-neuronal communication using binary-valued spikes.% and are more hardware friendly than their predecessors, i.e. ANNs. 
Significant research has been done on implementing SNNs on FPGA and Application Specific Integrated Circuits (ASIC). Mu\~noz et al. \cite{MUNOZ} implemented an SNN network using the Spike Response Model (SRM) and temporal coding on a Xilinx SPARTAN 3 FPGA to detect simple patterns. Wang et al. \cite{WANG} introduced a re-configurable, polychronous SNN with Spike Timing Dependent Delay Plasticity to fine-tune and add dynamics to the network. Time multiplexing was used to fit 4096 neurons and up to 1.15 million programmable delay axons on a VIRTEX 6 FPGA. Currently, most hardware SNN systems involve a preconfigured network that is implemented on an FPGA device to accelerate a specific task, leveraging the parallel processing capabilities of FPGAs. In other words, the parameters of the SNN, i.e., weights and thresholds of the neurons, are calculated using a simulator and are fixed in the hardware implementation.

Bethi et al. \cite{YESH} introduced an Optimized Deep Event-driven SNN Architecture (ODESA) that can be trained end-to-end using STDP-like rules which do not require continuous-valued gradients back-propagated. The ODESA training algorithm solves the credit assignment problem in SNNs by using the activity of the next layer in a network as a layer's supervisory signal. The synaptic weight adjustment in each layer only depends on the layer's trace and not on the weights of the other layers in the network. The feedback between the layers is causal and performed via binary event signals. The network does not require a symmetric backward pathway to perform training. This paper presents an efficient hardware implementation of a new SNN architecture utilizing the ODESA algorithm \cite{YESH}. Each layer has its training hardware module with minimal communication links with other layers. ODESA is an event-driven algorithm and has a very sparse activity due to the hard Winner-Takes-All (WTA) constraints on the layers. All the communication between the layers and the training modules is event-based and binary-valued. The ODESA architecture and its training algorithm provide an efficient, low-power, low-resource-consuming hardware implementation of SNNs that can be trained online and on-chip.

The remainder of the paper is organized as follows: Section \ref{sec:Background} reviews the background of Optimized Deep Event-driven Spiking Neural Network Architecture (ODESA). Section \ref{sec:Odesa_implementation} provides a detailed presentation of our heuristic SNN hardware implementation combined with its training hardware using the hardware-optimized ODESA algorithm. We will present two implementations of ODESA hardware experiments and their results. 
Finally, Section \ref{sec:Conclusion} presents the conclusion and directions for future works.

\section{Background}\label{sec:Background}

\begin{figure*}[h]
\centering
\includegraphics[width=0.85\textwidth]{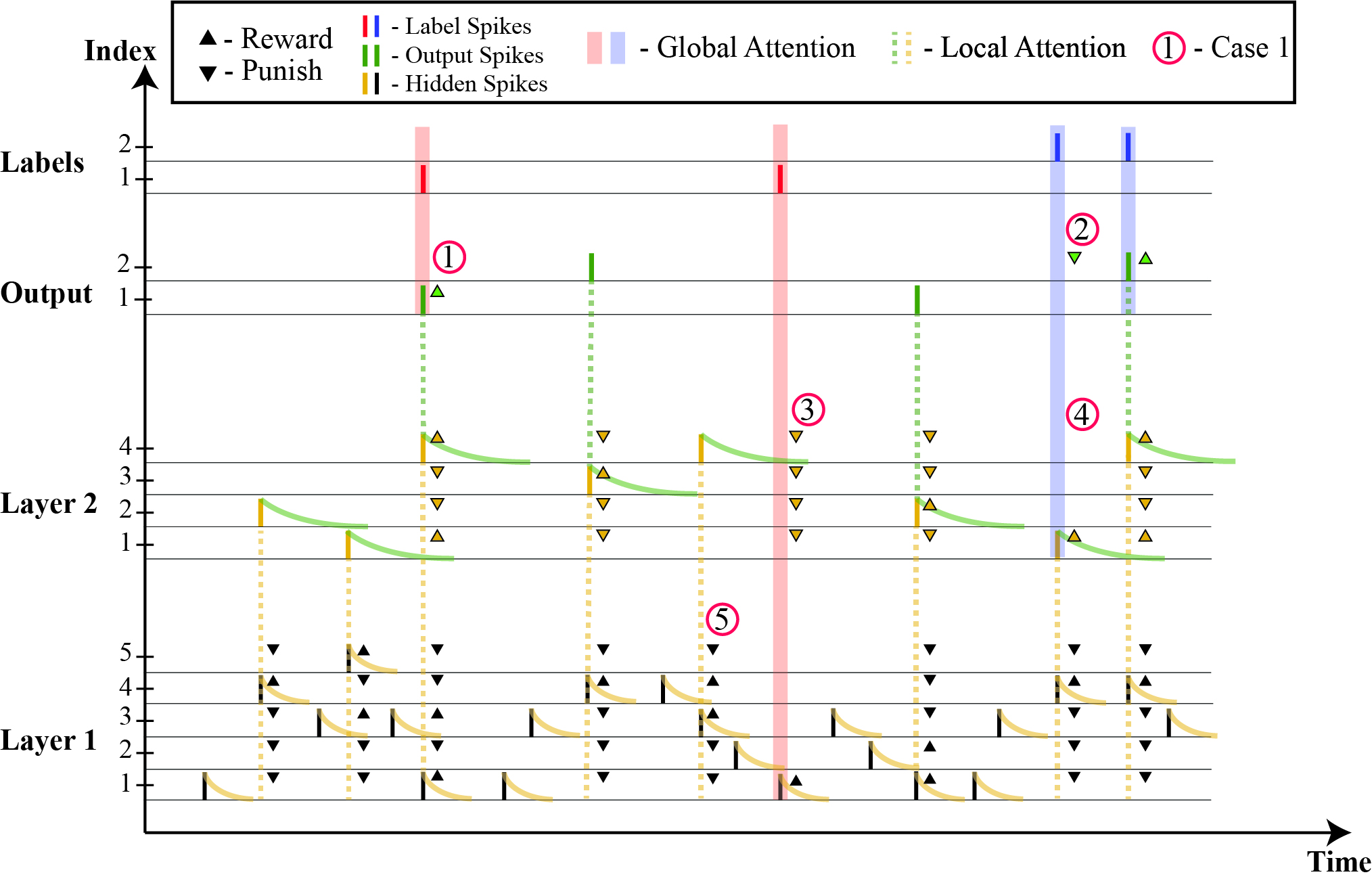}
\caption{Multi-Layer Supervision in ODESA using Spike-Timing-Dependent Threshold Adaptation. The shaded vertical lines represent the binary Global Attention Signal generated for each output label spike. The dotted vertical lines represent the binary Local Attention Signals sent to each layer from its next layer. The up and down arrows represent the reward and punishment of the individual neurons. Case 1: The predicted output spike matches the label spike, and the corresponding output neuron is rewarded. Case 2: The corresponding output neuron for the correct class is punished as it failed to spike in the presence of input from Layer 2. Case 3: All neurons in Layer 2 are punished as they failed to spike for an input spike from Layer 1 in the presence of the Global Attention Signal. Case 4: The active neuron in Layer 2 is rewarded in the presence of the Global Attention Signal. Case 5: The neurons with trace above the resent threshold are rewarded and the other neurons are punished in the presence of Local Attention Signal from Layer 2. Figure reproduced from \cite{YESH}.}
\label{fig:ODESASTDTP}

\end{figure*}
As the adoption of neuromorphic vision sensors increases, various dense tensor representations for sparse asynchronous event data have been investigated for learning spatio-temporal features in the data \cite{AFSHARINVEST}, \cite{MAQUEDA}, \cite{BALDWIN}.
A time surface, a term introduced by Lagorce et al. in \cite{LAGORCE}, is the trace of recent spiking activity at a given synapse at any time $t$.
The event-based time surface representations have been used in extracting features in tasks like space object detection and tracking \cite{COHEN}, \cite{AFSHARTRACK}, neuromorphic object recognition on UAVs \cite{ZAPPA}, and processing data from SPAD sensors \cite{AFSHARSPAD}. Afshar et al. \cite{AFSHAR} introduced an algorithm to extract features from event data using neuronal layers in an unsupervised manner called Feature Extraction using Adaptive Selection Thresholds (FEAST). FEAST is a highly abstracted and computationally optimized model of the SKAN method \cite{RACING}, \cite{TURN}. The FEAST method has been used and extended for a range of applications such as event-based object tracking \cite{RALPH}, activity-driven adaptation in SNNs \cite{HAESSIG}, and feature extraction to solve isolated spoken digits recognition task \cite{YINGTHESIS}, \cite{YING2022}.

In addition to weights representing the features learned by each FEAST neuron, each neuron has a threshold parameter that represents the size of the receptive field around the features represented by the weights. For every input event, the dot product of the time surface context and the synaptic weight vector of a neuron is calculated. 
The dot products of all neurons in a layer are compared to their respective thresholds. Only the neurons with dot products crossing their respective thresholds are eligible for selection. The neuron with the largest dot product in the eligible neurons is regarded as the winner for the given input event. If there is no winner, or in other terms, no neuron can cross its threshold, then the thresholds of all the neurons are reduced by a constant value. However, if a neuron becomes the winner, the weights of the neuron are updated with the current event context using an exponential moving average. The threshold of the winner neuron is also increased by a fixed value.

FEAST is an online learning algorithm that clusters the incoming event contexts of all the input events into clusters equal to the number of neurons used in the FEAST layer. The neurons' thresholds represent the clusters' boundaries (see Section 2.2 in \cite{AFSHAR}). Since there is no information about the significance of an individual event, FEAST treats each receiving event with equal priority, which results in learning features representing the most commonly observed spatio-temporal patterns in the input data. However, this may not be ideal for tasks that depend on more infrequent task-specific features. 

The Optimized Deep Event-driven Spiking neural network Architecture (ODESA) \cite{YESH} is a supervised training method that locally trains hierarchies of well-balanced Excitatory-Inhibitory (EI) networks on event-based data. ODESA is an extension and generalization of FEAST. The output classification layer in ODESA has $m \cdot N_c$ neurons ($m \in \mathbb{N}$) for a classification task with $N_c$ classes. The output layer is divided into $N_c$ groups (with $m$ neurons each), each responsible for one of the $N_c$ classes.
Each layer has a hard Winner-Takes-All (WTA) condition, which ensures only one neuron can fire in response to any input spike to a layer. The supervisory label spikes drive the threshold adaptation in an ODESA output layer for a given input spike stream. Since ODESA is event-driven, it is assumed that an input spike exists for every label spike. The labeled input spikes are treated with additional attention. For the labeled input spike, if there is no spike from the correct class neuron group, the thresholds for all the neurons in the class group are lowered. If there is a spike from any of the neurons in the correct class group, the winner neuron's weights are updated with the input spike's event context, and its threshold is also updated based on the dot product. Alternatively, in the absence of an output spike from the correct class group, thresholds of all neurons in the group are reduced. This weight update and threshold increase in a neuron can be considered ``rewarding a neuron" for its correct classification.
Similarly, a decrease in the threshold of a neuron to make it more receptive can be considered as ``punishing a neuron" for not being active. 

\begin{figure*}[ht]
  \begin{center}
   \begin{minipage}[b]{0.3\textwidth}
  \includegraphics[height= 2.5in]{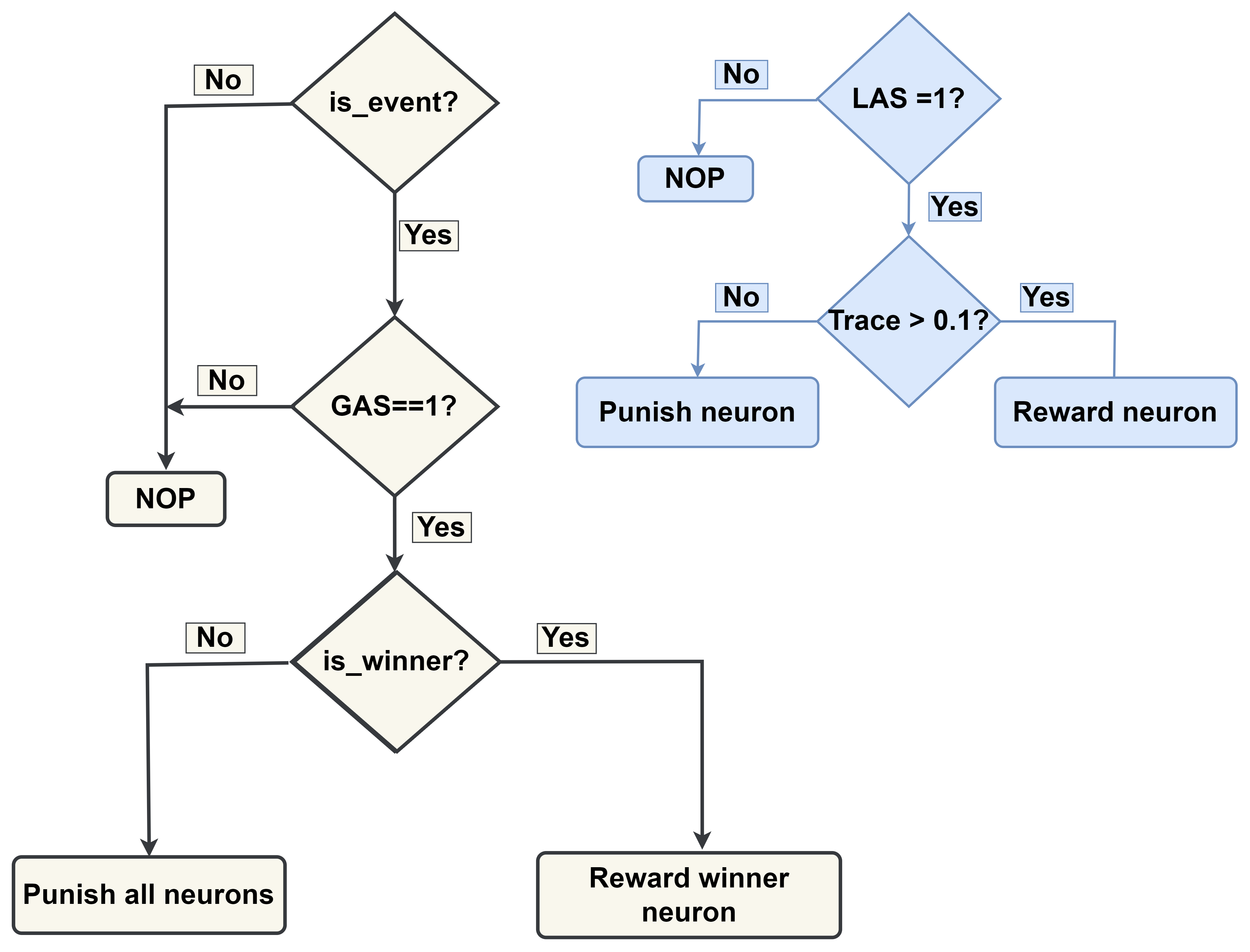}
  \caption{ODESA training algorithm for a hidden layer.} \label{fig:alg_h1}
  \end{minipage}
  \hfill
  \begin{minipage}[b]{0.5\textwidth}
    \includegraphics[height= 2.5in] {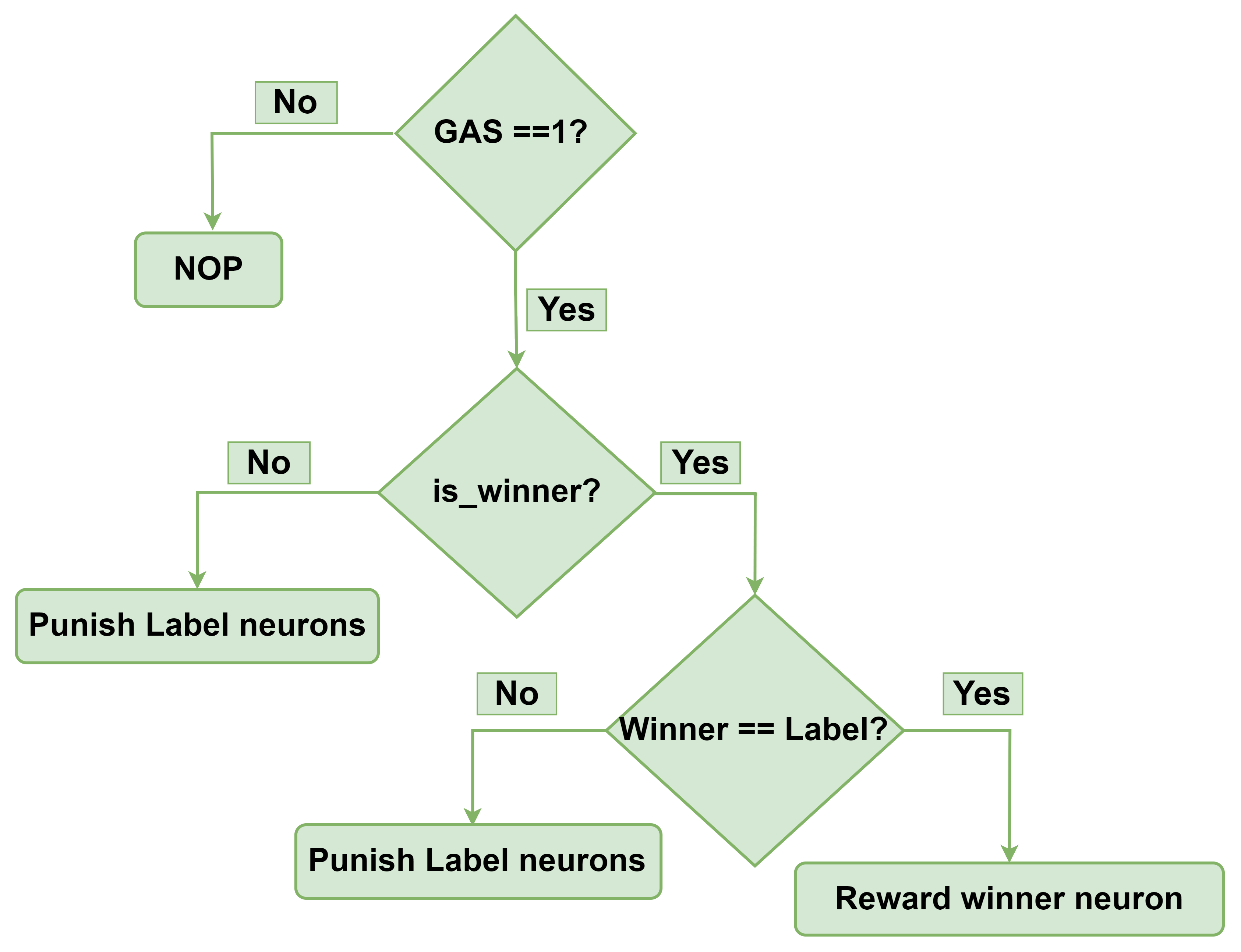} 
    \caption{ODESA training algorithm for an output layer}\label{fig:alg_h2}
  \end{minipage}
  \end{center}
\end{figure*}

The ODESA architecture can use multiple hidden layers with different time constants to learn hierarchical spatio-temporal features simultaneously at different timescales \cite{YESH}. Each hidden layer goes through a similar threshold adaptation as the output layer based on the spiking activity of its next layer in the hierarchy. A binary attention signal is generated by each layer to its previous layer whenever a neuron in the layer is active. All the neurons which were recently active in the previous layer are rewarded and the rest of the neurons are punished. These binary signals called Local Attention Signals (LAS), help provide the necessary feedback required to train the hidden layers. This architecture is well suited for enabling online learning in hardware as the communication between layers is through binary attention signals only, and there is no need to calculate loss functions and pass continuous-valued gradients across the layers during training. 
A Global Attention Signal (GAS) is generated when a label is assigned to an input spike. The GAS is accessible by all layers. Each layer also has access to the LAS generated by its next layer in the hierarchy. There is no LAS for the output layer. The output layer compares the generated spikes with the labels to reward or punish activated neurons. Fig. \ref{fig:ODESASTDTP} depicts the multi-layer supervision of ODESA architecture. The condensed ODESA algorithm is depicted in Fig.\ref{fig:alg_h1} and Fig. \ref{fig:alg_h2} flowcharts.

%figure 1,2

When a LAS is active, the training algorithm determines the participation of a neuron in generating a spike in the next layer based on its eligibility trace. If the trace of a neuron is above a certain limit (generally set to $10\%$ of its full scale), it is rewarded. The neurons with traces lower than the limit are punished. 

\section{ODESA hardware implementation} \label{sec:Odesa_implementation}

\subsection {Primitive building blocks of the ODESA network}
In this Section, we introduce the primitive building blocks of the ODESA network. The primitive building blocks are reusable in different ODESA network architectures.

\subsubsection{Synchronizer}
\begin{figure}[!ht]
  \begin{center}
  \includegraphics[width= 2in]{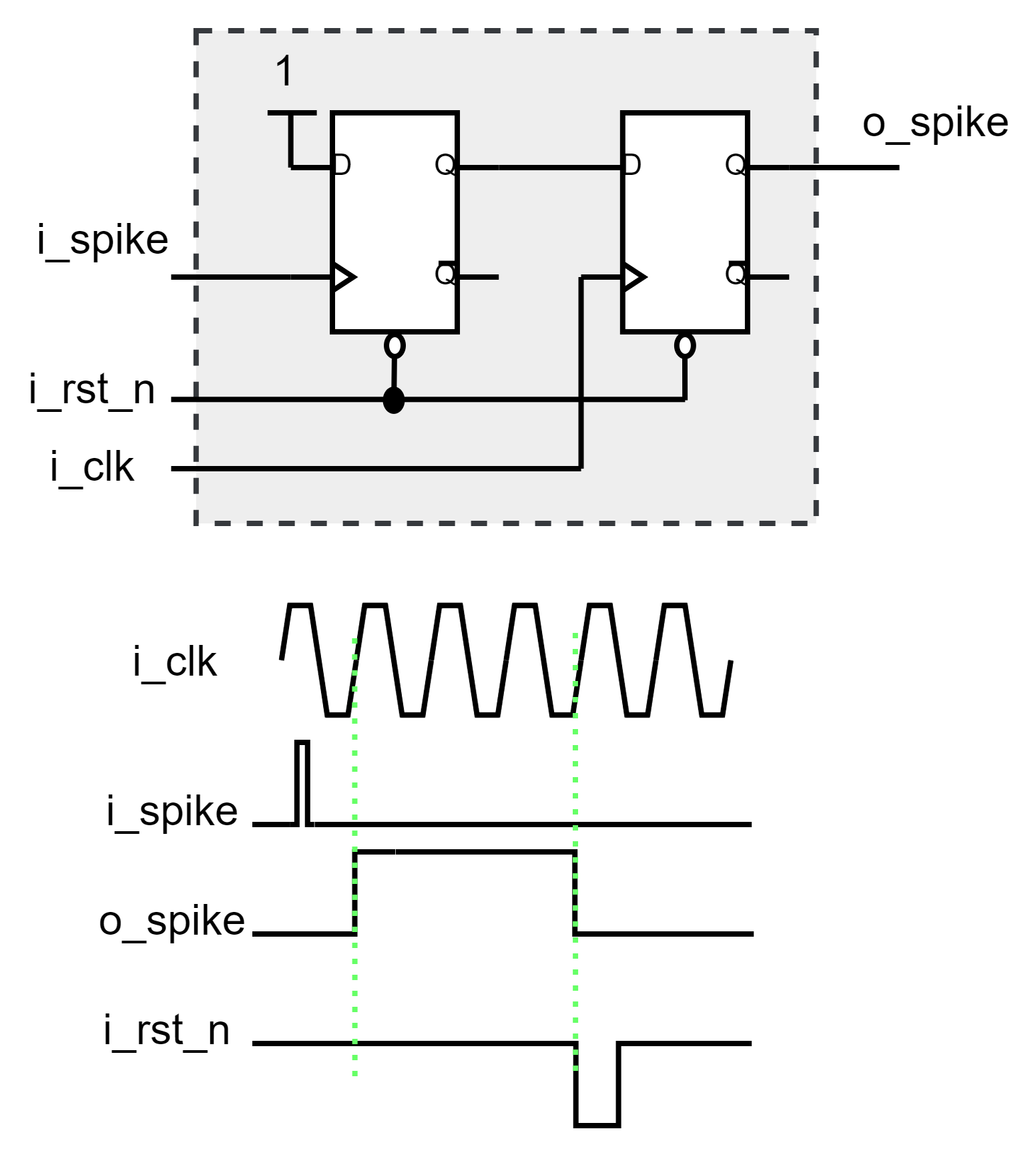}
  %\vspace{-15pt}
  \caption{Synchronizer module and its timing diagram. The `i\_spike' signal will be synchronized with the rising edge of `i\_clk'. If `i\_rst\_n' is not activated new spike will be ignored.}\label{fig:synchroniser}
  \end{center}
\end{figure}

The Synchronizer is used to synchronize the asynchronous input events (spikes) with the system clock. The input spikes to the ODESA network are not necessarily synchronous with the system clock and can be missed. Fig.\ref{fig:synchroniser} shows the design of a Synchronizer module. 
If an event happens at the input of the Synchronizer, the output will be asserted at the rising edge of the next clock. The Synchronizer will not respond to new events until it is reset by a logic through its `i\_rst\_n' input signal. This will let the system have control over accepting or rejecting events.

% ==== FIG 3

\subsubsection{Leaky accumulator} 
The Leaky accumulator is a modified digital implementation of the Leaky Integrate and Fire (LIF) model of a neuron \cite{LIF}. It is used to model the synaptic response to an incoming spike in the ODESA network. In our modified design, the leaky accumulator can model either linear or exponential decay. The linear decay accumulator consists of a base adjustable bit-width down-counter. Here, if an event/spike is received, the counter is reloaded to the output value of the adder and starts to count down at the rising edge of the following clocks until it decays to zero. Equation \ref{eq:eq_1} shows the value of the counter in the linear decay accumulator at time $t$ given a spike at time $t'$. Thus for a spike $\delta(t-t')$, which arrived at any time $t'$ between two consecutive clock cycles with time period $T$ (i.e. $(k-1) T < t' \leq k T \mid k \in \mathbb{N}$), the counter value of the Leaky accumulator, $a(t)$, at time $t= nT$, $n \in [k, C+k]$ can be expressed as:
\\
\begin{equation}
a(t) = (C-\frac{t-kT}{T})(u(t-kT)-u(t-(C+k)T),
\label{eq:eq_1}
\end{equation}
\\
where $C$ is the linear decaying constant that will be loaded to the counter when a spike happens and $u(t)$ is the unit step function. The decay rate is controlled by either the value of the decaying constant $C$ or the clock frequency. If a new event happens when the decaying counter is not zero, it will be reloaded by the sum of the current counter value and the constant $C$. For a stimulus $\delta(t-t_1)+\delta(t-t_2)$, where two spikes $\delta(t-t_1)$ and $\delta(t-t_2)$ occur close to each other at times $t_1$, and $t_2$ respectively, such that $t_1 < t_2 \leq t_1+ CT$, the counter value is calculated as: 
\\
\begin{equation}
a(t) = a(t_1) + a(t_2).
\label{eq:eq_2}
\end{equation}

\begin{figure*}[ht]
  \begin{center}
   \begin{minipage}[b]{0.4\textwidth}
  \includegraphics[width= 2.5in]{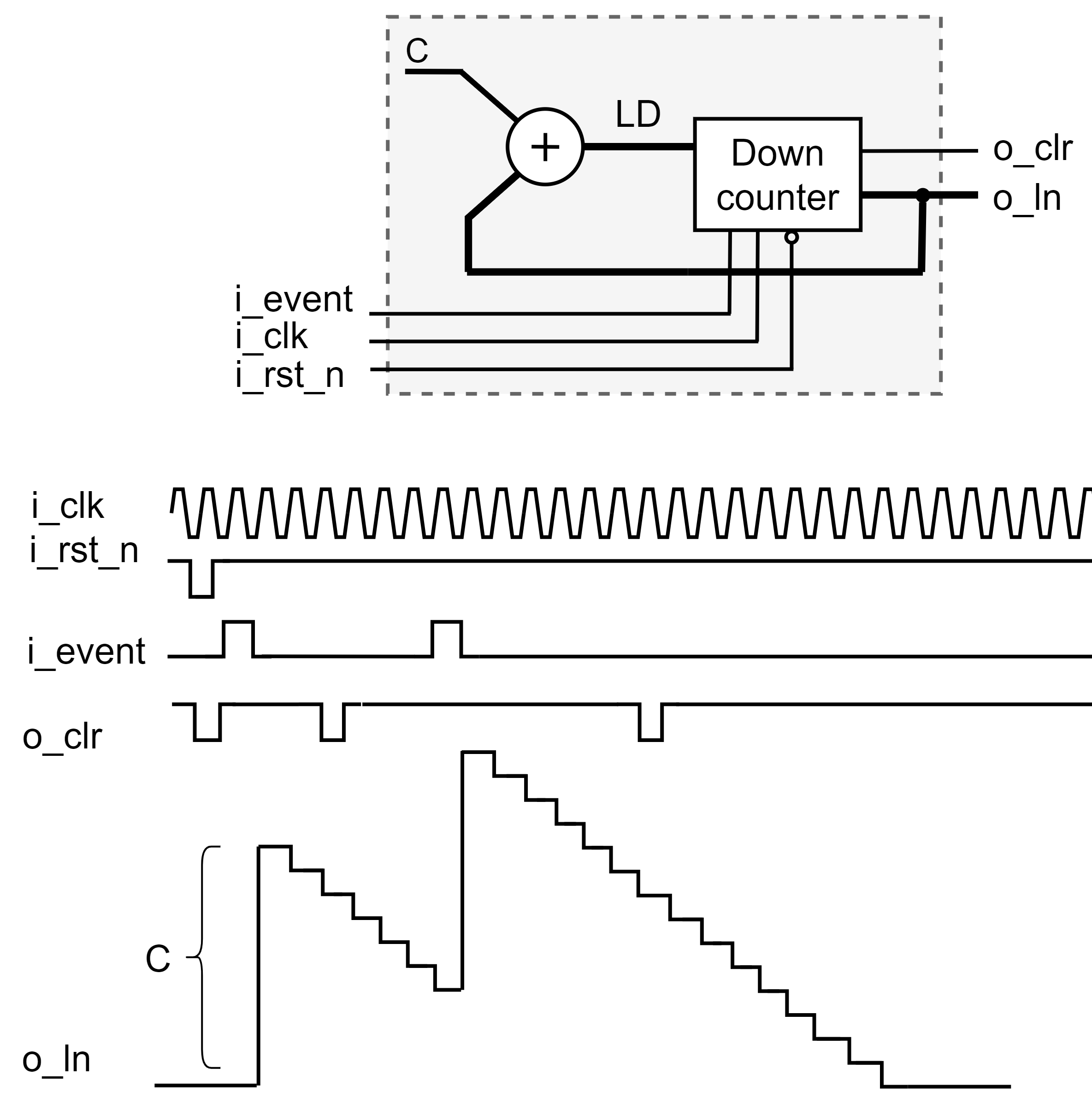}
  \caption{Leaky Accumulator architecture with linear decay and the circuit timing diagram with two subsequent input spikes.} \label{fig:lin_accu}
  \end{minipage}
  \hfill
  \begin{minipage}[b]{0.4\textwidth}
    \includegraphics[width= 2.5in] {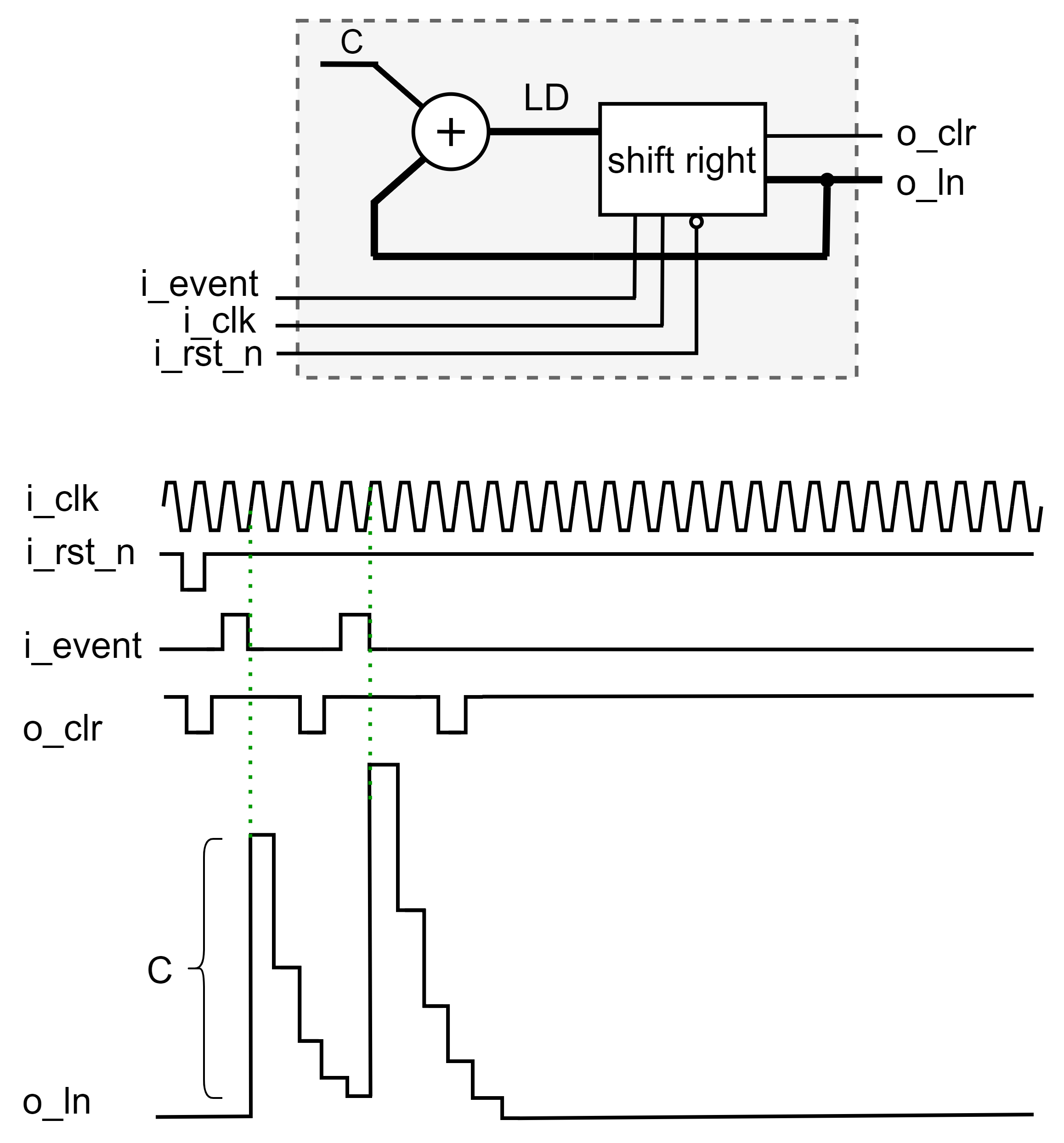} 
    \caption{Leaky Accumulator architecture with exponential decay and the circuit timing diagram with two subsequent input spikes.}\label{fig:exp_accu}
  \end{minipage}
  \end{center}
\end{figure*}

Fig. \ref{fig:lin_accu} shows the block diagram of a linear decaying Leaky accumulator and a sample waveform. The Leaky accumulator activates a clear signal (`o\_clr') three clock cycles after receiving a synchronized event. This signal is used to reset the Synchronizer and make it ready to receive new input events. The exponential decay is estimated by a divide by two (a shift right) at each clock cycle after loading constant $C$ into the shift register. The output of an exponential decaying Leaky accumulator, $a(t)$, will be:
\\
\begin{equation}
a(t) = {\frac{C}{2^{\frac{t}{T}-k}}} \left( u(t-kT)-u(t-(\tau+k)T) \right).
\label{eq:eq_3}
\end{equation}

For the exponential decay, $C$ is set to  $2^{\tau} -1$, where $\tau$ is the decay constant. In this work, we used linear decay accumulators only. Fig. \ref{fig:exp_accu} shows the architecture of an exponential decaying leaky accumulator.

% ==== FIG 4,5

\subsubsection{Synapse}
\begin{figure}[ht]
  \begin{center}
  \includegraphics[width=3.4in]{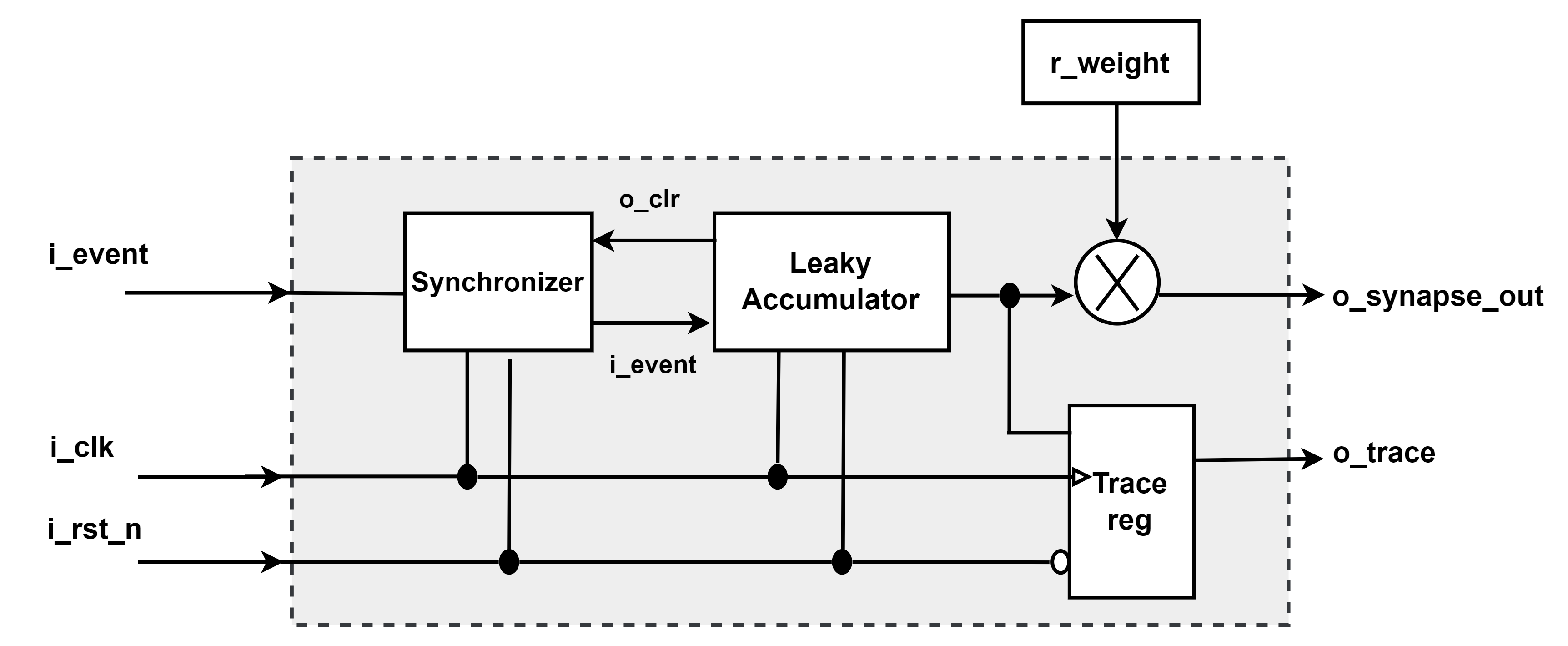}
 % \vspace{-15pt}
  \caption{Synapse architecture}\label{fig:synapse}
  \end{center}
\end{figure}

The Synapse module consists of a leaky accumulator and a weight multiplier. Capturing an asynchronous event forces the leaky accumulator to generate a decaying output amplified by the weight multiplier. The Synapse weight is stored in a register (`r\_weight'), and its value is determined during the network training process. The `r\_weight' register resides in the Training hardware, which will be detailed in Section \ref{sec:training_hardware}. The output of a Synapse $i$, $b_i(t)$, will be: 
\\
\begin{equation}
b_i(t) = {w_i}\cdot a(t),
\label{eq:eq_4}
\end{equation}
\\
where $w_i$ is the value saved in the `r\_weight' register of the Synapse $i$.
The `TRACE' register contains the time surface value (the output of the leaky accumulator) at every clock cycle. This value is used for training the previous network layer if it exists. Fig. ~\ref{fig:synapse} illustrates the architecture of a Synapse.

% ==== FIG 6

\subsubsection{Neuron}
Each Neuron comprises several Synapses. The outputs of all Neuron Synapses are added together. The resulting value is equivalent to the dot product calculated in ODESA \cite{YESH} and it is referred to as ``membrane potential" throughout this paper as used in a LIF neuron model. The membrane potential is compared with the Threshold register value. The output of the Neuron is the membrane potential value if it exceeds the Threshold register value; otherwise, it is set to zero. 
The Threshold register is also located in the training hardware, and its value will be assigned during the training phase. 
The output of a Neuron with $m$ Synapses can be written as: 
\\
\begin{equation}
d(t) = \begin{cases}
\sum_{i=1}^{m} b_i(t) & \text{if } d(t) \geq \text{Threshold}\\
0, & \text{otherwise.}
\end{cases}
\label{eq:eq_5}
\end{equation}

In an ODESA layer, the comparator and prioritizing module compare the output values of the neurons. The neuron with the highest membrane potential and the lowest index in the layer is declared as the winner. Subsequently, the comparator module generates a spike corresponding to the index of the winning neuron.
The membrane potential of the winner Neuron is latched in its `LAST\_VALUE' register. The `LAST\_VALUE' register is used during training. We will discuss the training process in detail in Section \ref{sec:training_hardware}.

\begin{figure}
  \begin{center}
  \includegraphics[width=3.2in]{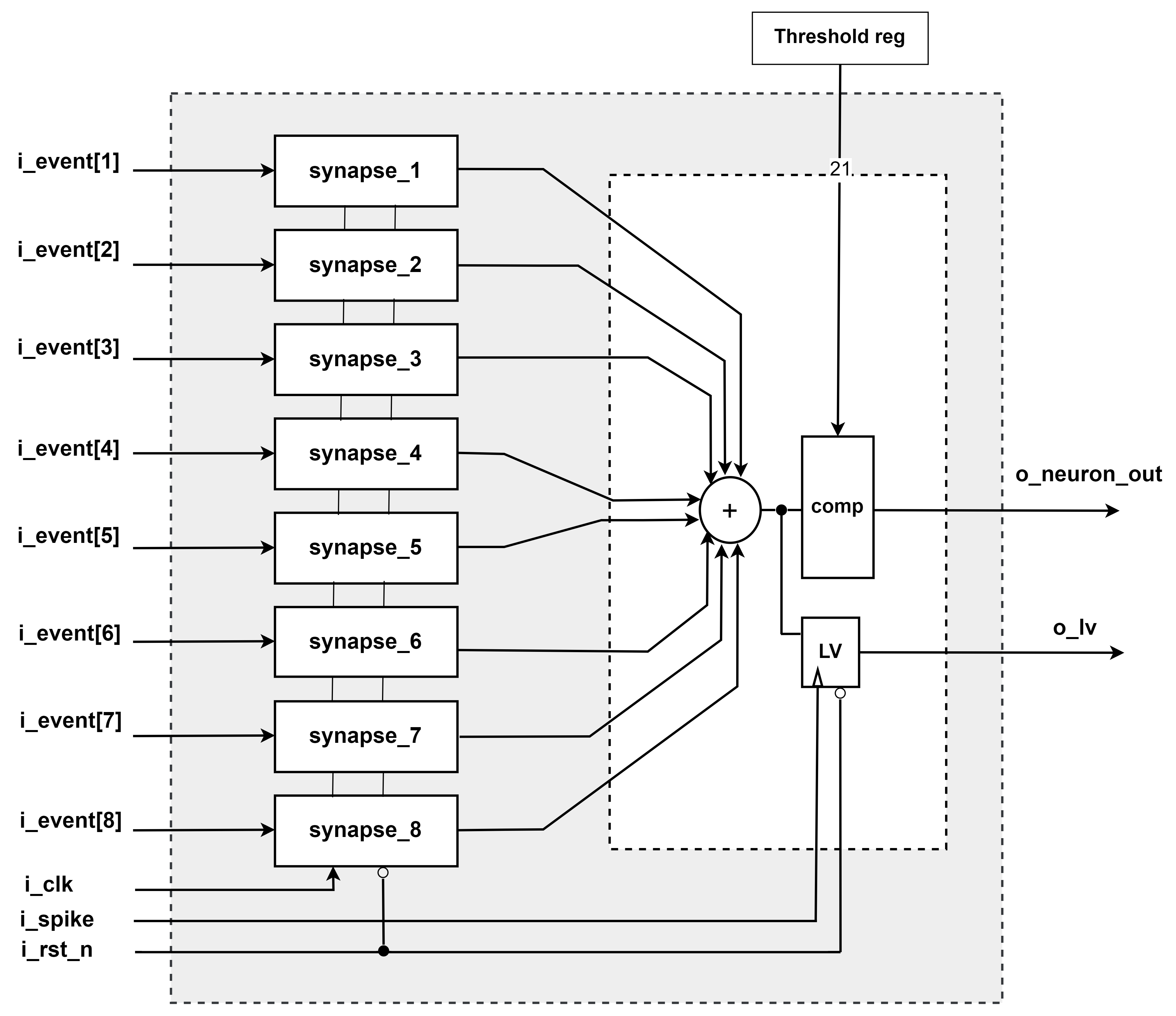}
 % \vspace{-15pt}
  \caption{8-input ODESA Neuron}\label{fig:neuron}
  \end{center}
\end{figure}

Fig. \ref{fig:neuron} depicts an 8-input neuron block diagram. The `i\_spike' input of the Neuron receives feedback from the output spike (`o\_neuron\_out'). The membrane potential is latched at the rising edge of the `i\_spike' input and can be accessed via the `o\_lv' output of the Neuron. 

% ==== FIG 6

\subsubsection{Comparator and Spike generator}
\begin{figure}[!ht]
  \begin{center}
  \includegraphics[width=3.2in]{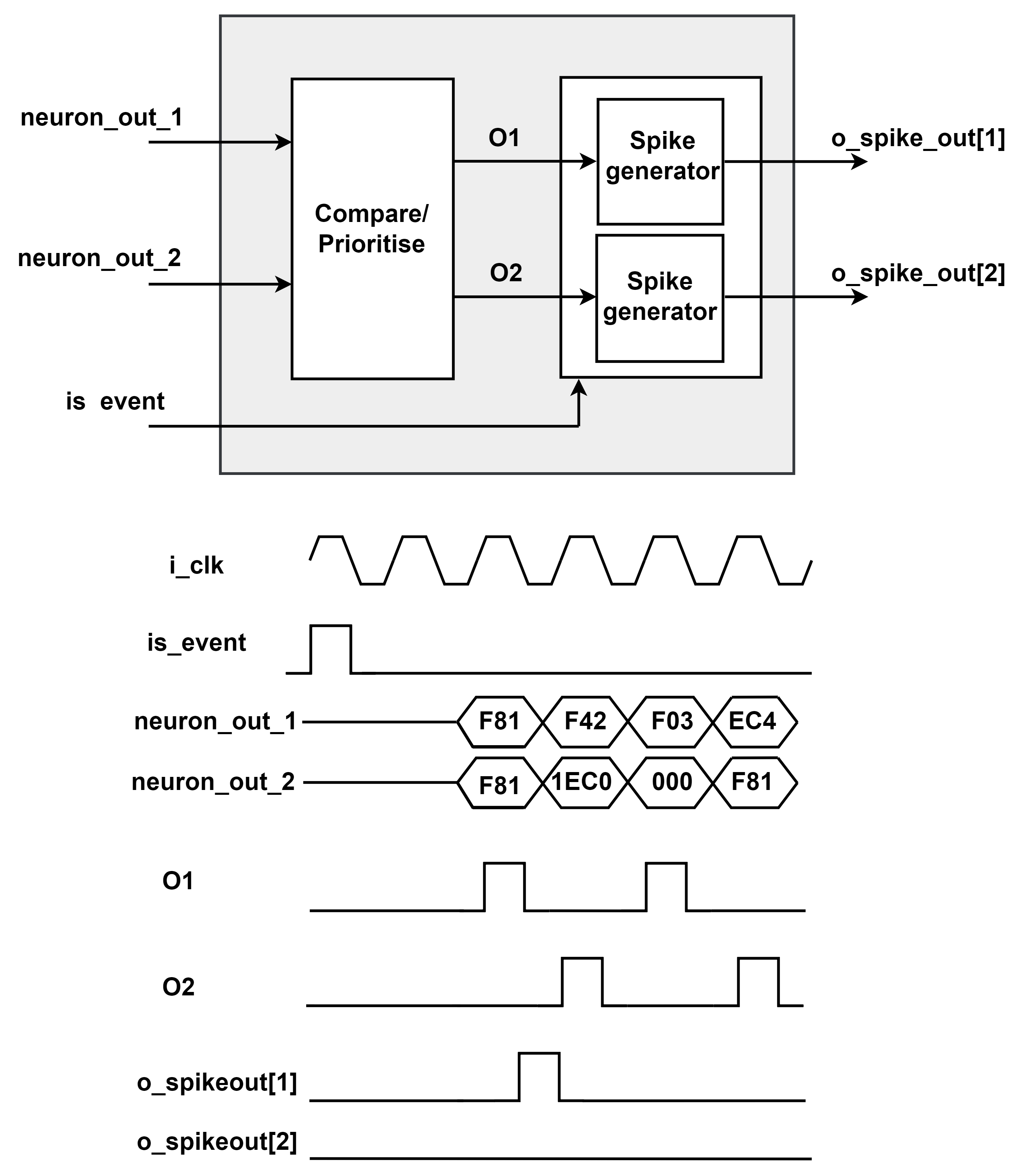}
  %  \vspace{-15pt}
  \caption{2-input Comparator and Spike generator and sample waveform}\label{fig:comparator}
  \end{center}
\end{figure}

\begin{figure}[!ht]
  \begin{center}
  \includegraphics[width=3.2in]{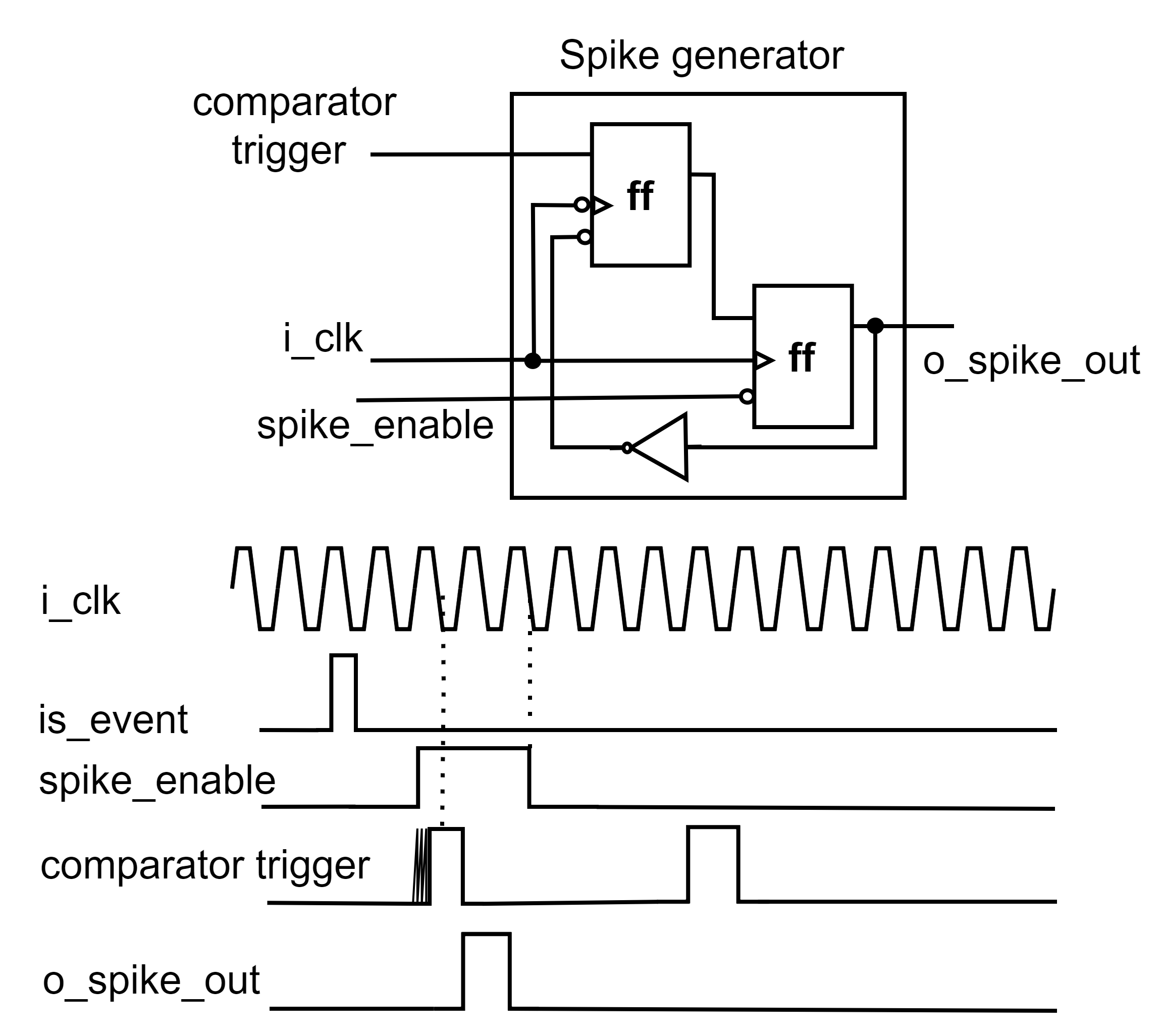}
  %  \vspace{-15pt}
  \caption{Spike generator module and sample waveform}\label{fig:spike_gen}
  \end{center}
\end{figure}

The Neurons' outputs are received by the Comparator module that detects which Neuron has the higher membrane potential and prioritizes the neuron outputs based on the input index to the Comparator module. The lower the index, the higher the priority of the Neuron. The Comparator output for an $n$ neuron layer, is the post-synaptic spike stream of all the Neurons in a layer, and it can be mathematically modeled as:
\\
\begin{equation}
\resizebox{.5 \textwidth} {!}{
$e_i(t) =
\begin{cases} 
\delta(t), &
\text{if:}
\begin{cases}
    & {\text{IS\_EVENT} = 1},\\
    & d_{i-1}(t)< d_i(t)\ge \{d_{i+1}(t),\ldots,d_n(t)\}.\\
\end{cases}\\

0, & \text{otherwise,}
\end{cases}$
}
\end{equation}
\\
where $i$ is the Neuron index in an ODESA layer with $n$ Neurons and the `IS\_EVENT' signal indicates whether any input event has occurred during the recent clock cycles.
The Comparator output is one-hot encoded indicating the winner Neuron (the one with the highest membrane potential and the lowest index). Due to the event-driven computation, the Comparator must have an output only if the Neuron becomes a winner due to an input event. This is critical to avoid generating unwanted or unrelated spikes at the output of the ODESA layer and removing unintended spurs generated by the Comparator's combinatorial logic, which can cause intermediate spikes even when there is no input spike to the layer. The Spike generator is a sequential logic that receives a Comparator's output and allows a spike to appear at the output if an input event is recorded a few clock cycles before. The number of clock cycles that the Spike generator module can look back on is adjustable for each module. In our design, after detecting an input event, the spike generator waits until four clock cycles to receive a signal from the Comparator module. Fig. \ref{fig:comparator} shows the block diagram and the function of the Comparator. Fig. \ref{fig:spike_gen} illustrates the Spike generator logic and a sample waveform.

% ==== FIG 8

% ==== FIG 9

\subsubsection{ODESA SNN Layers}
All ODESA layers, either an input, a hidden, or an output layer, have a homogeneous architecture. That is, a number of Neurons are connected to a Comparator module. Neurons can have different numbers of Synapses. However, every Neuron has only one output. The number of layer outputs is equal to the number of Neurons in the layer. As discussed, only one of the layer outputs can be active at any given time. Neurons within a layer share the inputs to the layer. The outputs of a layer are fully connected to the inputs of the following layer, except for the output layer whose outputs indicate the classes in the classification problem the ODESA Network is designed to solve. Normally, different ODESA layers operate at different clock periods. The ratios of the hidden and output layers' clock period to the input layer's clock period are part of the network's configuration parameters. 

\begin{figure*}[ht]
  \begin{center}
  \includegraphics[width=5in]{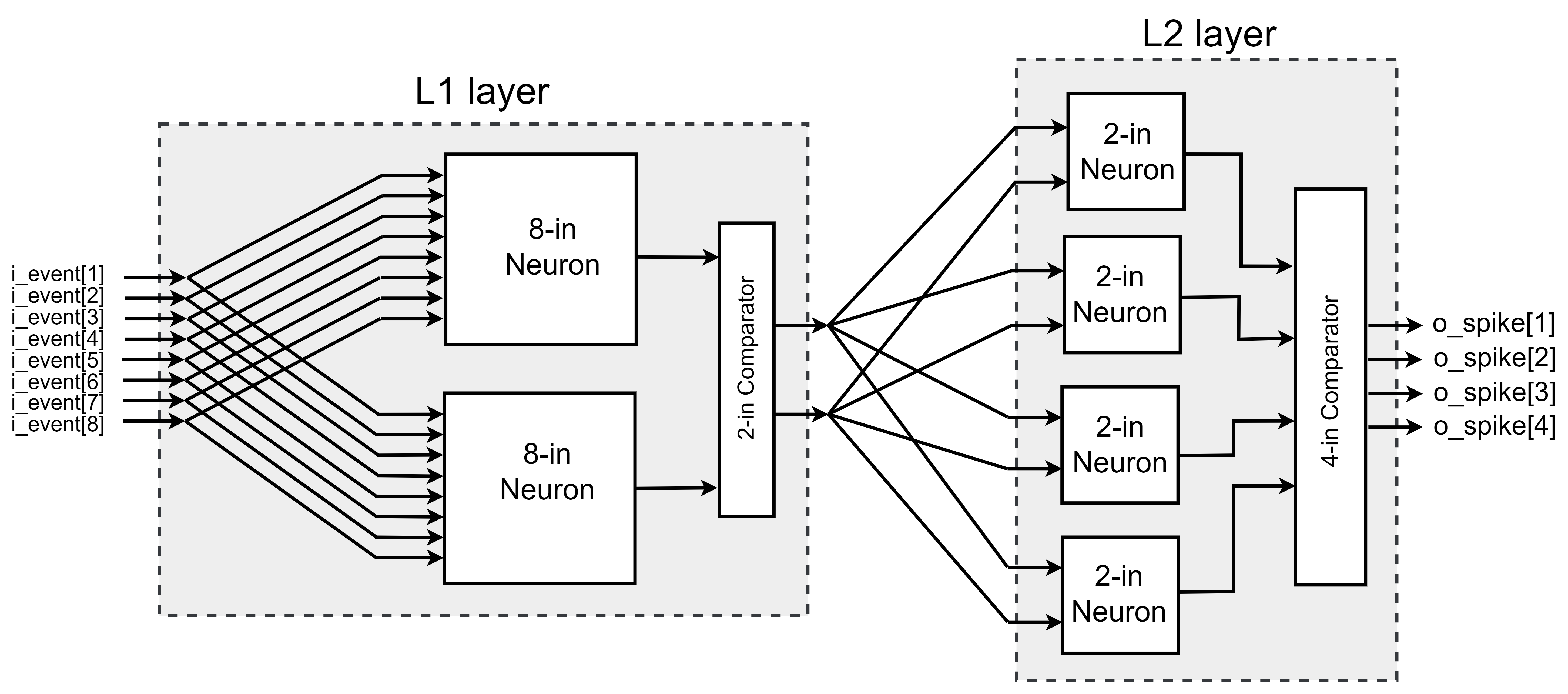}\\
  \caption{Two layer ODESA implementation. Layer one (L1) is the input layer with two 8-input neurons. Layer 2 (L2) is the output layer with four 2-input neurons that classify the inputs into four classes.} \label{fig:circuit_diagram}
  \end{center}
\end{figure*}

We use a naming convention throughout this paper to reference the architecture of the ODESA network. 
The input layer is always called `L1'. Then, we increment the  Layer's number for the following layers to the output layer, e.g. `L2', `L3', and so forth. Any ODESA network architecture can be determined using the following naming convention: ODESA \{number of input spike channels\}\_\_\{number of neurons at level 1 \}\_\textellipsis\_\{number of neurons at level n\}\_\_\{number of output classes\}. Fig. \ref{fig:circuit_diagram} shows an example of ODESA 8\_\_2\_4\_\_4 architecture.

\section{Training hardware} \label{sec:training_hardware}
\begin{figure*}[!ht]
  \begin{center}
  \includegraphics[width=5in]{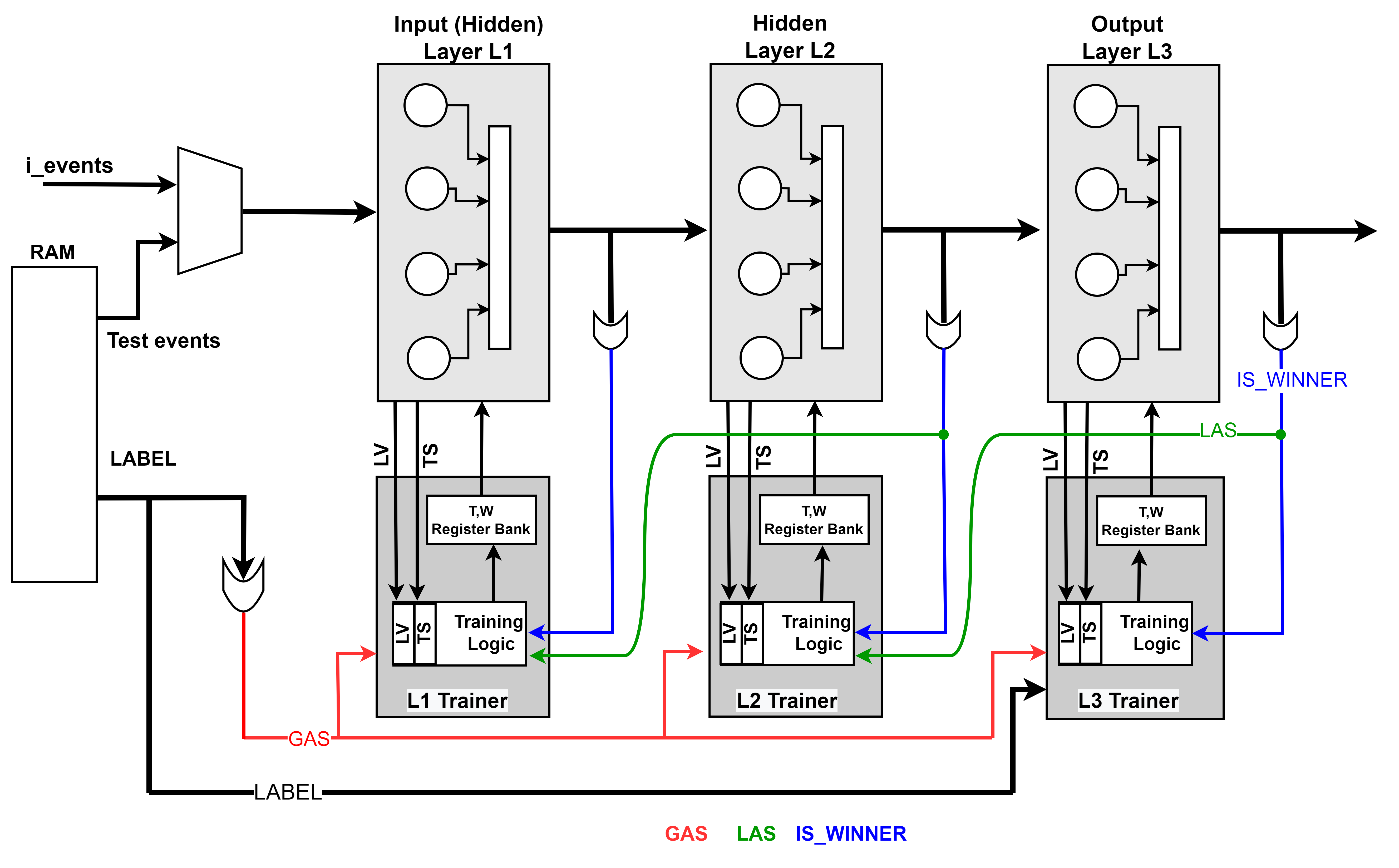}
  %  \vspace{-15pt}
  \caption{ODESA network, Neuron layers, training hardware, and connections.}\label{fig:training_layers}
  \end{center}
\end{figure*}

ODESA is a multi-layer supervised Spiking Neural Network architecture that can be trained to map an input spatio-temporal spike pattern to an output spatio-temporal spike pattern without requiring access to the weights and thresholds of other neurons or batching of the input data. 
The training algorithm is distinct for hidden layers (including the input layer) and the output layer. At each layer, the training is done through the guiding signals produced by the successive layer, the layer's output spikes, and the Label spikes. The original algorithm is detailed and implemented in software in \cite{YESH}. In this work, we represent a revised version of the algorithm enhanced for hardware implementation. 
If any layer fires a spike, an `IS\_WINNER' signal is generated for that layer's training logic. Each layer's training logic also receives a Local Attention Signal (LAS) and a Global Attention Signal (GAS).
If there is a spike at the output of a layer, a LAS signal will be generated for its preceding layer. The GAS signal, however, is generated when a label spike exists for the current input spike and propagates through all layers. The training set which includes the input spikes and their corresponding labels is stored in the RAM. During the training phase, the input spikes are read from the RAM and injected into the input layer of the ODESA SNN. Likewise, the training hardware reads labels from the RAM and compares them with the output spikes generated by the output layer. Fig. \ref{fig:training_layers} illustrates an ODESA network with the network layers and training logic for each layer. 

The training hardware for each layer receives the last value of the membrane potential and the trace of all Synapses in that layer.
When a Neuron becomes a winner (the `IS\_WINNER' signal is asserted), a (post-synaptic) spike is generated at the output of the layer, and the value of the Synapses' Trace registers (Fig. \ref{fig:synapse}), are latched in a time surface (TS) register. The value of the membrane potential (the adder output in Fig. \ref{fig:neuron}) is also registered in the Last Value (LV) register. The TS register is implemented in the Training module (not visible in Fig. \ref{fig:training_layers} for simplicity) and its value represents the contribution of the Synapse to the generation of the winner's membrane potential. If the Neuron remains silent in presence of an input event and GAS signal, then the trace of the Synapse is latched to the `NO\_WINNER' register that is implemented in the Training module. The high value of the `NO\_WINNER' register indicates that the layer failed to spike for an input spike to the layer. 

The update of weights and thresholds happens in the presence of the GAS signal in a ``reward" or ``punish" process. ``reward" or ``punish"  is a similar process for all neurons across the layers. For a Neuron $j$ with threshold $T_j$, and $s$ Synapses, the reward process is defined according to Equation \ref{eq:eq_7}.
\\
\begin{equation}
\begin{aligned}
Reward:
\begin{cases}
w_{ij} &\leftarrow  w_{ij} + {\eta}_{w} \cdot  (TS_{ij} -  w_{ij}), \forall i \in [1,s],\\
T_{j} &\leftarrow T_{j} + {\eta}_{T} \cdot (LV_{j} - T_{j}),
\end{cases} 
\end{aligned}
\label{eq:eq_7}
\end{equation}
\\
where $w_{ij}$ is the synaptic weight, and $TS_{ij}$ is the time surface register of Synapse $i$ of Neuron $j$. $LV_{j}$ is the Last Value register of Neuron $j$. 
The Neuron ``punish" process is just lowering Neuron's threshold by the constant value $\Delta_T$ as stated in Equation \ref{eq:eq_8}.  
\\
\begin{equation}
Punish: \;
T_{j}  \leftarrow T_{j} -{\Delta_T},
\label{eq:eq_8}
\end{equation}
\\
where ${\eta}_{w} < 1$, and ${\eta}_{T} < 1$, are the learning rates of the layer, and ${\Delta_T} \geq 1$ are the network hyper-parameters. For sake of a low-cost hardware implementation, learning rate values are chosen as negative powers of two; therefore, the ``reward"  can be performed by simple shift and addition operations. Usually, the ${\eta}_{w}$ and ${\eta}_{T}$ are set to the same value. 

Since the Weight and Threshold registers contain unsigned integer values, special consideration has to be taken to ensure that the product terms of ${\eta}_{w}$, and  ${\eta}_{T}$ in Equation \ref{eq:eq_7} will never become zero, leaving the training process in a locked state.  Additionally, when experiments require learning rates that are too small to perform weight updates via shift operations, the Weight and Threshold update steps are reduced to Equation \ref{eq:parameterizedUpdates}. The $\operatorname{sign}<>$ function is used to determine the direction of the weight (or threshold) change and is then updated by a fixed step equal to $\eta_w$ (or $\eta_T$). $\eta_w$ and $\eta_T$ values are set to the lowest possible step changes (e.g. 1,2,3,..) like used in Section \ref{sec:iris_dataset}    
\\
\begin{equation}
\begin{aligned}
Reward:
\begin{cases}
w_{ij} &\leftarrow  w_{ij} + {\eta}_{w} \cdot \operatorname{sign}  (TS_{ij} -  w_{ij}), \forall i \in [1,s],\\
T_{j} &\leftarrow T_{j} + {\eta}_{T} \cdot \operatorname{sign} (LV_{j} - T_{j}),
\end{cases} 
\end{aligned}
\label{eq:parameterizedUpdates}
\end{equation}
\\
Algorithm \ref{alg:HLT} and \ref{alg:OLT} show the hardware-friendly ODESA training algorithms for the Hidden and Output layers, respectively.  

\begin{algorithm}[ht]
\caption{Training algorithm for Hidden Layers.}\label{alg:HLT}
\SetKwInput{Input}{Input}\SetKwInOut{Output}{Output}
\Input{$IS\_WINNER,\; LAS,\; GAS$}
%\Output{$T$, $W$ update}
\BlankLine
%\emph{$i=0$}\;
\uIf{(IS\_WINNER \& GAS)}
{
Reward the WINNER Neuron;
}
\uIf{(!IS\_WINNER \& GAS)}
{
Punish All Neurons;
}
\uIf{(LAS)}
{
\ForEach{Neuron}
{
\uIf{TRACE $>$ 10\% FS)}
{
Reward Neuron;
}
\uElseIf
{NO\_WINNER $>$ 10\% FS)}
{
Punish Neuron;
}
}
}
\BlankLine
\emph{return Updated \{$T$, $W$\} }
\end{algorithm}
\begin{algorithm}[ht]
\caption{Training algorithm for output Layer. \\}\label{alg:OLT}
\SetKwInput{Input}{Input}
\SetKwInOut{Output}{Output}
\Input{$GAS, WINNER, LABEL$}
\SetKwSwitch{Case}{When}{Other}{case}{do}{when}{otherwise}
%\Output{Updated \{$T$, $W$ \} }
\BlankLine

\uIf{($GAS$)}
{
%\uIf{(WINNER == LABEL)}
%{Reward LABEL Neuron;}
%\uElseIf{(WINNER==0)}
%{Punish LABEL Neuron;}
%\uElseIf{(WINNER != LABEL)}
%{Negative weight update WINNER Neuron;
%Punish LABEL Neuron;}
%}
\Case{$WINNER$}
{
\When{$LABEL$}
{
Reward $LABEL$ Neuron;
}
\When{$0$}
{
Punish $LABEL$ Neuron;
}
\Other{
Negative Weight update $WINNER$ Neuron;\\
Punish $LABEL$ Neuron;
}
}
}
\emph{return Updated \{$T$, $W$\} }\\
\end{algorithm}

\begin{figure}[ht]
  \begin{center}
  \includegraphics[width=3.0in]{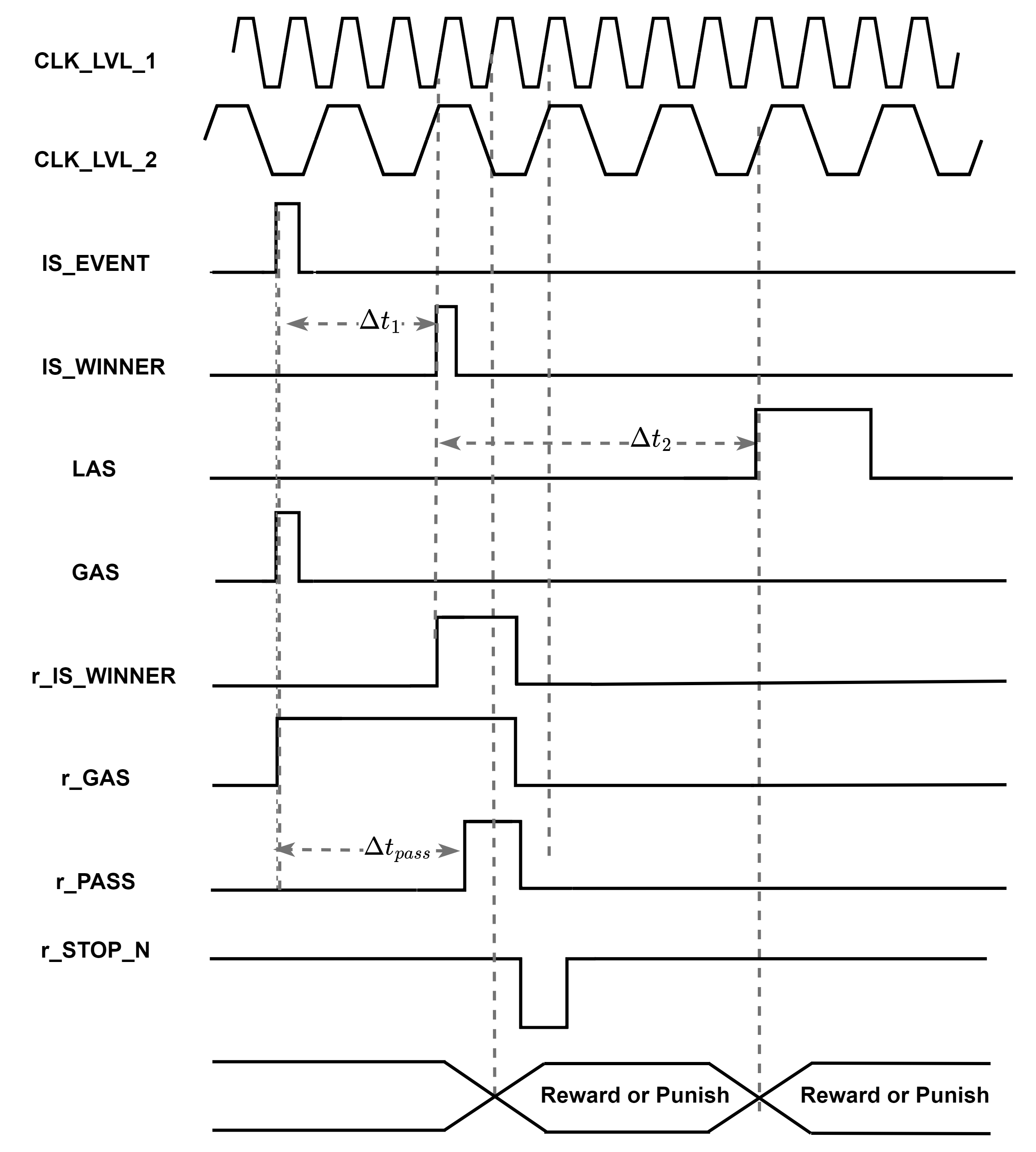}
  %  \vspace{-15pt}
  \caption{ODESA Hidden layer training signals.} \label{fig:training_wf1}
  \end{center}
\end{figure}

Since the `IS\_WINNER' and GAS signals have no overlap, we use a latched version of these signals in the hardware. Fig. \ref{fig:training_wf1} shows the training waveforms for a hidden layer. The GAS signal is asserted at the same time `IS\_EVENT' becomes active (events and labels are read simultaneously from the RAM). The `IS\_WINNER' spikes after $\Delta t_{1}$ time passed from `IS\_EVENT'. The `r\_IS\_WINNER', and `r\_GAS' signals are latched and verified after $\Delta t_{pass}$ time on the rising edge of the ODESA level's clock. The LAS signal is asserted after $\Delta t_{2}$ time from `IS\_WINNER'. The `IS\_WINNER' also indicates there exists an input event for the next layer. The values of $\Delta t_{1}$, $\Delta t_{2}$, and $\Delta t_{pass}$ are configurable at the Neuron's architecture. Specifically, $\Delta t_{i}$ represents the time required for a spike to appear at the output of ODESA layer $i$ following any input event to the layer. On the other hand, $\Delta t_{pass}$ represents the time that the training module waits to observe the winner and update the weights and thresholds. In our design, all of these parameters are set to 3 clock cycles of the corresponding layer's clock.
For an output layer, however, the WINNER is compared with the LABEL in the event of a GAS signal. If the WINNER and LABEL match, then the winner Neuron is rewarded; otherwise, the winner Neuron's weights are suppressed by a negative weight update. The negative weight update is considered the reverse of the weight reward process, i.e., 
\\
\begin{equation}
w_{ij} \leftarrow  w_{ij} +{\eta}_{w} \cdot  w_{ij} - {\eta}_{w} \cdot TS_{ij} \;\; \forall i \in[1,s]. 
\label{eq:eq_9}
\end{equation}

\begin{figure}[ht]
  \begin{center}
  \includegraphics[width=3.0in]{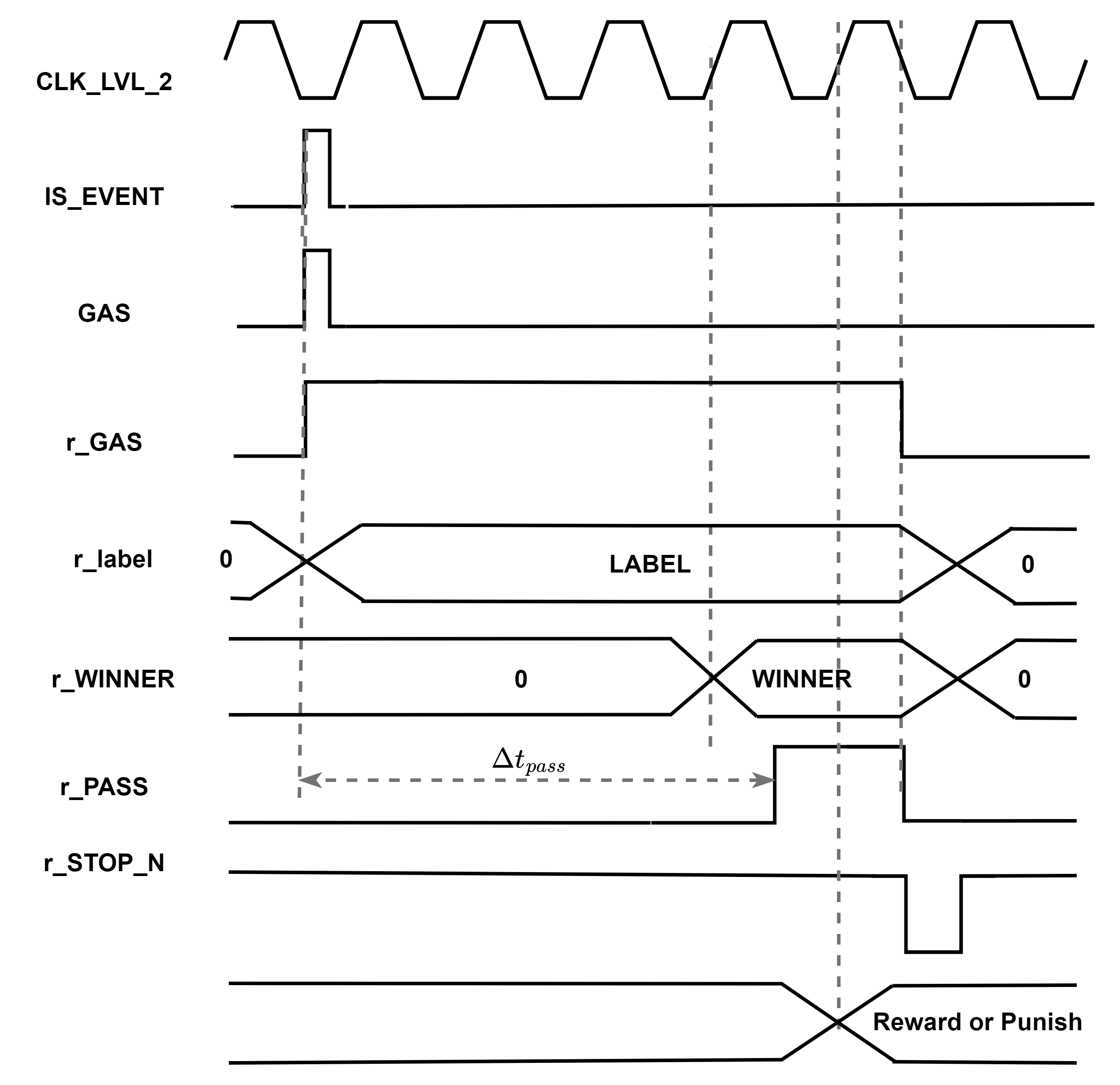}
     % \vspace{-5pt}
  \caption{ODESA Output Layer training signals.}\label{fig:training_wf2}
  \end{center}
\end{figure}

Fig. \ref{fig:training_wf2} demonstrates the training signals for the output layer. The LABEL is latched at the rising edge of the GAS signal. It takes $\Delta t_{pass}$ time for the WINNER to appear at the output layer, which is then compared with the LABEL at the rising edge of the next clock and performs weight and thresholds update.

% === FIG 13

\section{ODESA network Experiments}

\subsection{Experiment 1, Detection of four classes of spike patterns}
\begin{figure*}[ht]
  \begin{center}
  \includegraphics[width=6.5in]{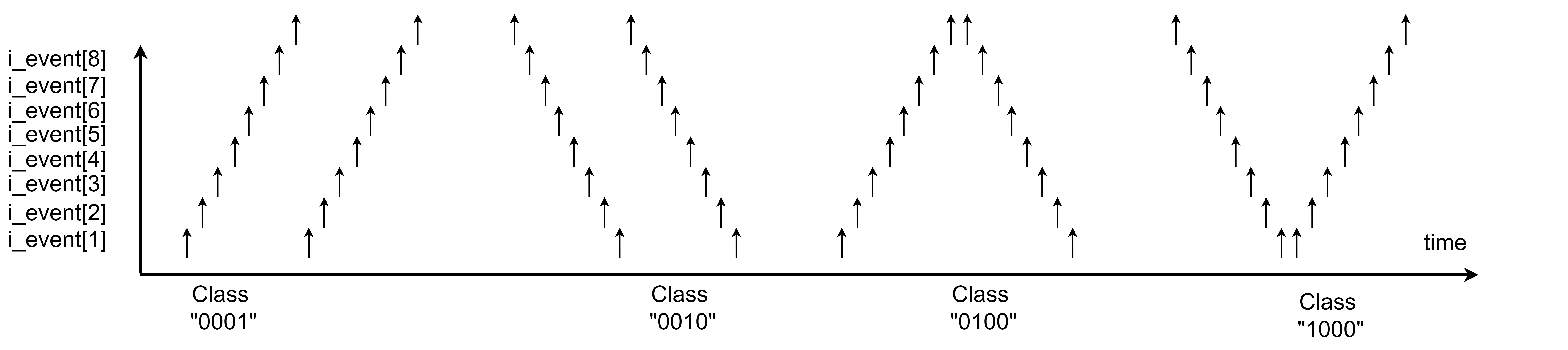}
  %  \vspace{-15pt}
  \caption{Experiment 1 input spike patterns.}\label{fig:slash_pat}
  \end{center}
\end{figure*}

Our first experiment uses ODESA to detect four patterns consisting of 16 spikes split into two sub-patterns of 8 spikes each, which appear at a uniform time distance of  $\nu$. A label spike was attached to the last spike of each of the four patterns. The four input event patterns can be mathematically expressed in Equation \ref{eq:exp_1}. Fig. \ref{fig:slash_pat} visualizes the four spike patterns in time and the assigned class label for each pattern.
\\
$\forall i \in [1,8]: $
\begin{equation}
\begin{aligned}\resizebox{.5 \textwidth} {!}{$
    \begin{cases}
        Pattern\; 1: i\_event[i]= \delta(t-(i-1)\nu) + \delta(t-(8+i)\nu)\\
        Pattern\; 2: i\_event[i]= \delta(t-(9-i)\nu) + \delta(t-(17-i)\nu)\\
        Pattern\; 3: i\_event[i]= \delta(t-(i-1)\nu) + \delta(t-(17-i)\nu)\\
        Pattern\; 4: i\_event[i]= \delta(t-(9-i)\nu) + \delta(t-(8+i)\nu)\\
    \end{cases} $}
\end{aligned}
\label{eq:exp_1}
\end{equation}

The ODESA network implemented for this application is configured with two fully connected layers. The input layer (L1) has two Neurons with eight inputs, and the Output layer (L2) has four Neurons with two inputs. The network architecture is illustrated in Fig.\ref{fig:circuit_diagram}. 
The Network parameters used for detecting the four class patterns are listed in Table \ref{tab:slash_par}. 

\begin{table}[ht]\small{
  \begin{center}
    \caption{ODESA 8\_\_2\_4\_\_4 parameters for experiment 1}
    \label{tab:slash_par}
    \begin{tabular}{l l l } %{|c|c|c|}
    \hline 
    \emph{Parameter} & \emph{L1} & \emph{L2}  \\  
     \hline 
     ${\eta}_w $  & $2^{-3}$ &  $2^{-2}$  \\ 
     \hline 
     ${\eta}_T $  & $2^{-3}$ &  $2^{-2}$  \\ 
     \hline 
     ${\Delta_T} $  & $2^{6}-1$ &  $2^{6}-1$ \\ 
     \hline 
     \emph{Weight register (bits)} & 8 & 8 \\    
     \hline 
     \emph{Decaying counter (bits)} & 6 & 6 \\    
     \hline     
     \emph{Clock frequency (MHz)}  &  0.064 &  0.032 \\  
    \hline 
    \emph{Input events time distance $\nu$ (ms)} & 31.25  &  - \\ 
    \hline 
    \end{tabular}
  \end{center}
  }
\end{table}

In this experiment, we use 6-bit linear decaying counters at each Neuron and a decaying constant $C = 63$. The clock frequency for L1  and L2 is set to 64 kHz and 32 kHz, respectively. Thus, the linear decay to zero will take one millisecond for Neurons at L1 and two milliseconds for Neurons at L2. The distance between two spikes is set to $\nu = 8 \times \text{(L1 clock period)}$.

\begin{figure*}[ht]
  \begin{center}
  \begin{minipage}[b]{0.45\textwidth}
  \includegraphics[width=3.1in]{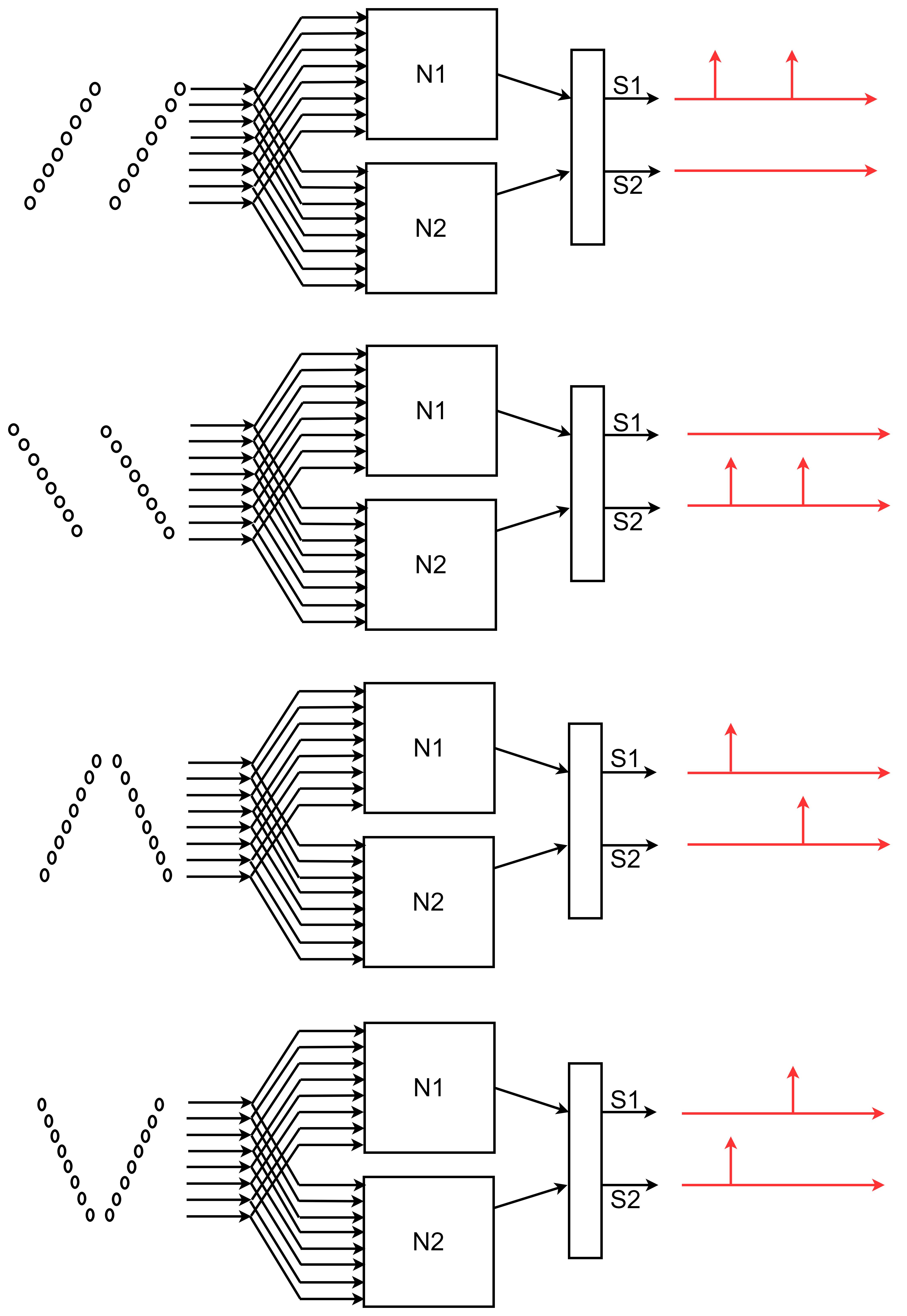}
      \vspace{3pt}
  \caption{ODESA Layer L1 response to input events.}\label{fig:L1}    
  \end{minipage}
  \hfill
 \begin{minipage}[b]{0.45\textwidth}
  \includegraphics[width=3.2in]{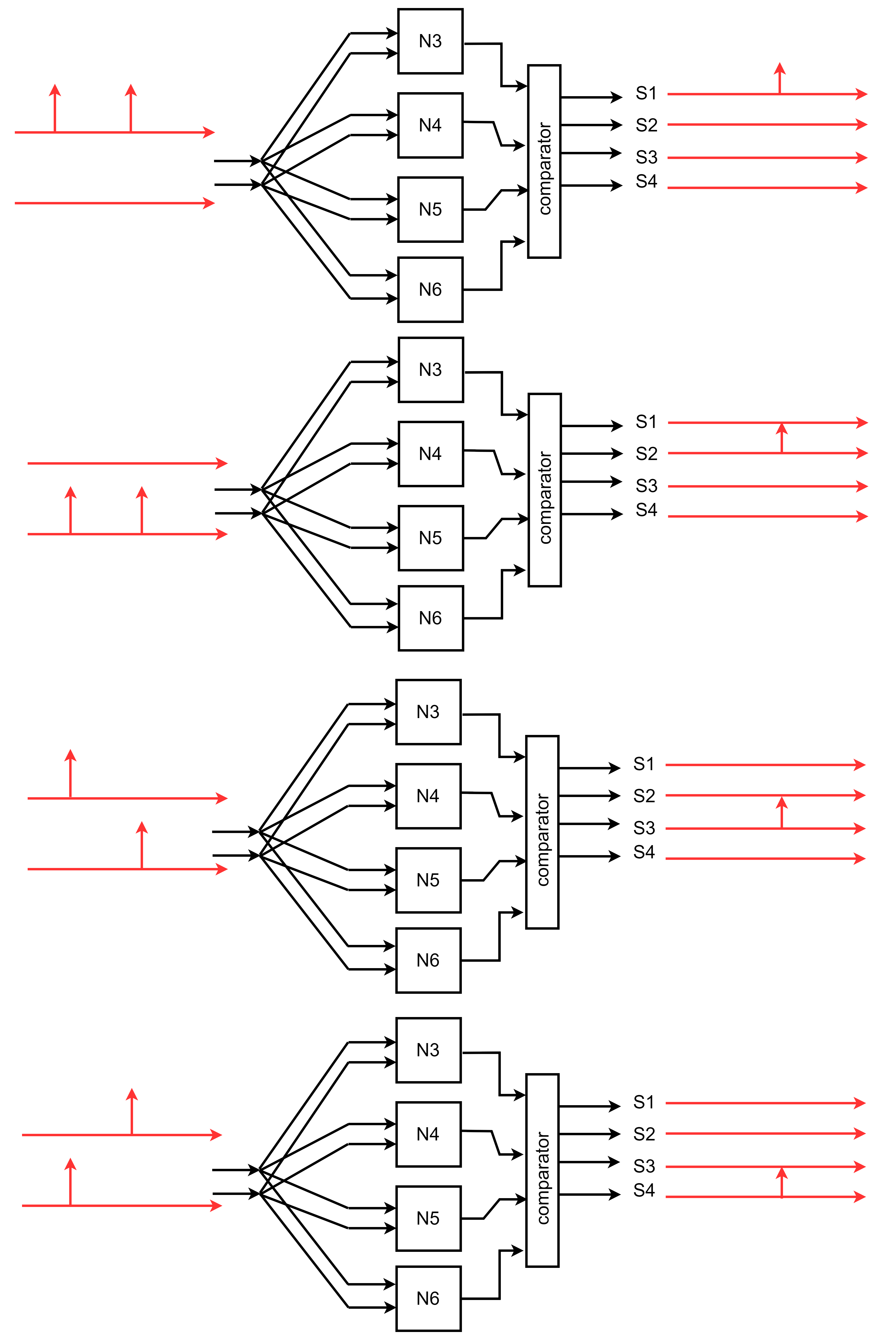}
     % \vspace{-5pt}
  \caption{ODESA Layer L2 response to input events.}\label{fig:L2} 
 \end{minipage}
  \end{center}
\end{figure*}

As shown in Fig. \ref{fig:L1}, for each pattern, two spikes are generated by L1. The position and time of the spike determine the input pattern injected into the ODESA network. The L1 output spikes are used as inputs to L2. As depicted in Fig. \ref{fig:L2}, L2 comprises four 2-input neurons that perform the classification task.

The ODESA 8\_\_2\_4\_\_4 implementation was performed on an Intel Cyclone V (part no. 5CSEBA6U23I7) using the Quartus 18.0 Lite design tool. Table \ref{tab:exp_1_fpga} reports the implementation results. 

The implemented ODESA network achieved an accuracy rate of 100\% after completing self-training. The accuracy did not change by applying random changes in the distance $\nu$ by $\pm 10\%$ on a trained network.

\begin{table}[ht]\small{
  \begin{center}
    \caption{ODESA 8\_\_2\_4\_\_4 Implementation results on Intel Cyclone V }
    \label{tab:exp_1_fpga}
    \begin{tabular}{l l } %{|c|c|c|}
    \hline 
    \emph{Architecture} & \emph{ODESA 8\_\_2\_4\_\_4}  \\  
     \hline 
     \emph{Used ALM} &  1192  \\ 
     \hline 
     \emph{Used registers} &  976 \\ 
     \hline 
     \emph{Used DSP units} &  20 \\ 
     \hline 
     \emph{L1 max. Clock frequency (MHz)} & 28.25 \\ 
     \hline 
     \emph{Dynamic power consumption (mW)} & 1 \\      
    \hline 
    \end{tabular}
  \end{center}
  }
\end{table}

\subsection{Experiment 2, Iris dataset classification }
\label{sec:iris_dataset}
The Iris dataset \cite{FISHER} is one of the well-known databases in the pattern recognition literature for being not linearly separable. Different spike-encoding schemes are used to convert the Iris dataset into spikes to test the local learning rules of SNNs \cite{BOHTE}\cite{TAHERKHANI}\cite{SPOREA}. The data set contains three classes of 50 instances each, where each class refers to a type of iris flower. The four features are sepal length (d1), sepal width (d2), petal length (d3), and petal width (d4), all in centimeters within the range $[0.1, 7.7]$. To convert the input feature values into spatio-temporal spike patterns, we used a latency coding that maps the value of each input dimension to the time of a spike generated from a corresponding input channel:
\\
\begin{equation}
    L \to \delta(t- L).
    \label{eq:ltot}
\end{equation}

However, the length $L$ for $d_1$, $d_2$, $d_3$, and $d_4$ has to be scaled to fit in a fixed-length frame. In our case, that is the time frame within the range $[0,30]$. The dataset conditioning we applied to the original Iris dataset follows the offset and compress formulae in Equation \ref{eq:iris}.
\\
\begin{equation}
    \begin{cases}
         d_1 =\lceil 3.8 \big{(}\frac{(d_1 -1)}{2}+4 \big{)} \rceil,\\
         d_2 = \lceil3.8 \big{(}\frac{(d_2 -2)}{3}+2.5 \big{)} \rceil,\\          
         d_3 = \lceil 3.8 d_3 \rceil,\\
         d_4 = \lceil 9 (d_4 +0.5) \rceil.
    \end{cases}
\label{eq:iris}
\end{equation}

\begin{figure}[!ht]
  \begin{center}
  \includegraphics[width=3.4in]{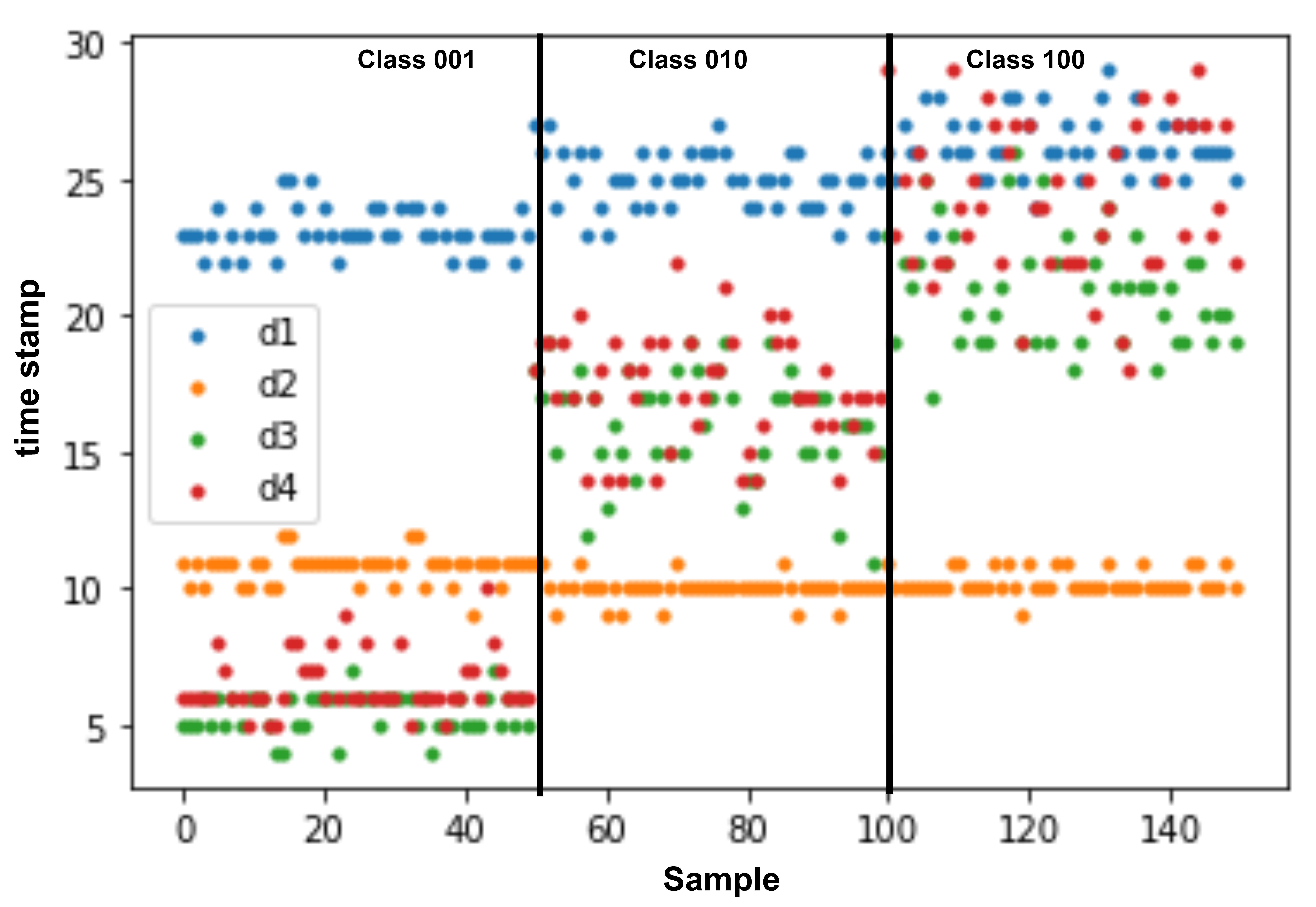}
  %  \vspace{-15pt}
  \caption{Experiment 2, Iris dataset input spike patterns.}\label{fig:iris_1}
  \end{center}
\end{figure}

Each sample in the new dataset contains the four features scaled to the timestamp ranged in $[0,30]$. Using Equation \ref{eq:ltot} the lengths $d1$, $d2$, $d3$, and $d4$  are converted to the timestamps. The dataset with timestamped events is shown in Fig. \ref{fig:iris_1}.  

The ODESA architecture we designed for the Iris dataset with four input spikes within the timeframe $[0,30]$ is ODESA 4\_\_6\_3\_\_3. The L1 includes 6 Neurons with 4 Synapses, and the L2 comprises 3 Neurons with 6 Synapses each. The clock frequency for L1 is 2.5 MHz, i.e., $\frac{1}{20}$ of FPGA system Clock (50 MHz). Clock frequency for L2 is set to 0.625 MHz, i.e., $\frac{1}{4}$ of L1 clock frequency. The ratio of the level one clock to the level two clock is a network parameter, which in this experiment is set to four. Samples are injected with the same clock frequency of L1. Therefore, each sample's timeframe takes a maximum of $30 \times 0.4 =12$ microseconds.

The decaying counter designed for this application is eight bits wide, and the decaying constant is set to its maximum value $C=255$ for both L1 and L2. 
The ODESA 4\_\_6\_3\_\_3 was implemented on Intel's Cyclone V FPGA, and the results are reported in Table \ref{tab:exp_2_fpga}.

\begin{table}[ht]\small{
  \begin{center}
    \caption{ODESA 4\_\_6\_3\_\_3 Implementation results on Intel Cyclone V }
    \label{tab:exp_2_fpga}
    \begin{tabular}{l l } %{|c|c|c|}
    \hline 
    \emph{Architecture} & \emph{ODESA 4\_\_6\_3\_\_3}  \\  
     \hline 
     \emph{Used ALM} &  2805  \\ 
     \hline 
     \emph{Used registers} &  1195 \\ 
     \hline 
     \emph{Used DSP units} &  42 \\ 
     \hline 
     \emph{L1 max. clock frequency (MHz)}  & 39.88 \\ 
     \hline 
     \emph{Dynamic Power consumption (mW)} &  $<$ 2 \\      
    \hline 
    \end{tabular}
  \end{center}
  }
\end{table}

\subsubsection{Training ODESA 4\_\_6\_3\_\_3 for Iris dataset}
Since the Iris dataset is more complex than our previous experiment with patterns of Fig. \ref{fig:slash_pat}, it requires smaller learning rates than that can be achieved by shift operations. For weights with smaller values, the shift operation can lead to no updates. We have used the weight and threshold update steps introduced in Equation \ref{eq:parameterizedUpdates}. The $\eta_w$ for L1 is set to 1. 
The resulting Weights register update for each Synapse $j$ of the Neuron $i$ follows the rule in Equation \ref{eq:wupdate}.
\\
\begin{equation}
\begin{cases}
     w_{ij} \leftarrow w_{ij} + 1, \text{if } TS_{ij} > w_{ij},\\
     w_{ij} \leftarrow w_{ij} - 1, \text{if } TS_{ij} < w_{ij}.   
\end{cases}
\label{eq:wupdate}
\end{equation}

The new weight update guarantees that the Synapse's weight value moves smoothly towards the time surface of that Synapse.
Rewarding the threshold is also performed by incrementing the threshold value by a fine-tuned constant value $\eta_T$ according to Equation \ref{eq:parameterizedUpdates}. This constant value $\eta_T$ is determined by trial. In our test, we set the $\eta_T$ equal to 127 decimals (or $0X7F$ hexadecimal). 

At L2, the weight updates require higher learning rates, and a larger step size is used for L2 to achieve the same. According to Equation \ref{eq:wupdate_l2}, each Synapse's weight is incremented or decremented by two in a rewarding process. 
\\
\begin{equation}
\begin{cases}
     w_{ij} \leftarrow w_{ij} + 2, \text{if }  TS_{ij} > w_{ij},\\
     w_{ij} \leftarrow w_{ij} - 2, \text{if }  TS_{ij} < w_{ij}.   
\end{cases}
\label{eq:wupdate_l2}
\end{equation}

The threshold update is done by employing Equation \ref{eq:eq_7} with $\eta_{T} = 2^{-10}$. The ``punish" process uses an adaptive $\Delta_T$ value according to Equation \ref{eq:adaptive_d} to ensure the Threshold register will never cross zero. 
\\
\begin{equation}
\Delta_T =
\begin{cases}
     2^{10} -1, \; \text{if }  T_j > 2^{16} -1,\\
     2^{8} -1, \; \text{if }  T_j > 2^{12} -1,\\
     2^{4} -1, \; \text{if }  T_j > 2^{8} -1,\\
     1, \; \text{otherwise. } \\
\end{cases}
\label{eq:adaptive_d}
\end{equation}
During our experiments, we noticed that training can be significantly accelerated by masking the LAS signals of the output layer, except for the ones that occur after a GAS signal. In other terms, generating LAS signals only when we anticipate a response from the network to the classification problem leads to fewer training epochs being necessary.

\begin{figure}[!ht]
  \begin{center}
  \includegraphics[width=3.4in]{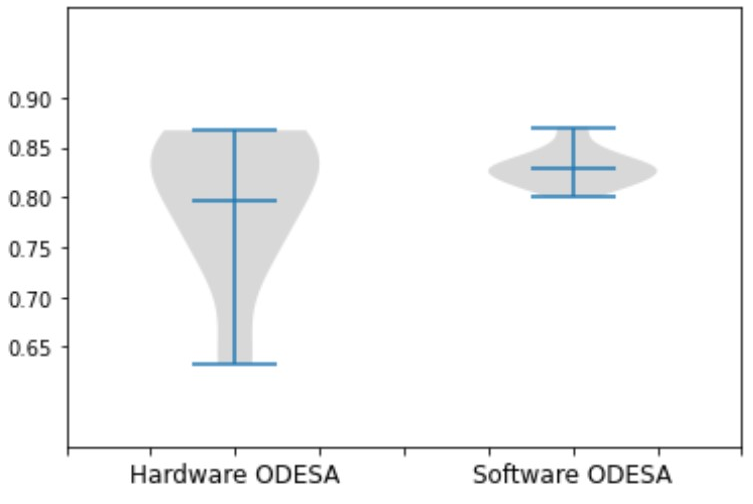}
  %  \vspace{-15pt}
  \caption{Comparison of the Accuracy in  Hardware and Software versions of ODESA, on the random splits of IRIS dataset}\label{fig:results}
  \end{center}
\end{figure}

To evaluate our network's performance and accuracy, we chose 20 random splits of the IRIS dataset (30\% training and 70\% test splits) and run the hardware with the training dataset splits stored in its RAM. 
The same dataset was used for the software version of ODESA with a similar ODESA 4\_\_6\_3\_\_3 configuration. We have used a smaller ODESA network with fewer input and hidden neurons compared to the network architecture (ODESA 20\_\_10\_3\_\_3) used in \cite{YESH} to fit the area available on Intel's Cyclone V. The input dimensions of the data are 4 as compared to 20 used in the original work \cite{YESH}. The dataset was converted to spikes using latency coding as described in Section \ref{sec:iris_dataset} as opposed to using a population code that was used in \cite{YESH} to reduce the number of multipliers required for the hardware implementation. The software version of ODESA is the original algorithm from \cite{YESH} that used floating-point operations and normalized weights and input time surface for calculating the dot products of the neurons. The dot product in the software version is always bounded between 0 and 1.  
%We also implemented a fully connected Artificial Neural Network (ANN) using Keras back-end to compare our hardware ODESA network performance with an ANN. The ANN consists of four neurons at the input level with a ReLu activation function, three neurons at the output level with a softmax activation function, a categorical\_crossentropy loss function, and an Adam optimizer. 
We let the two networks (our ODESA hardware, and software ODESA) run for 400 epochs on each random split. The accuracy performance of the networks is abridged in Fig. \ref{fig:results}. The average and maximum achieved accuracy and the standard deviations are reported in Table \ref{tab:iris_compare}. The Software ODESA shows a consistent accuracy with a small variation of around 83\%, while the hardware version accuracy changes from a maximum of 86.6\% to 65\%. %The ANN accuracy has the highest variation and changes in the range of 58\% to 97.7\%.

% \begin{table*}[ht]\small{
%   \begin{center}
%     \caption{Iris dataset Accuracy test comparison }
%     \label{tab:iris_compare}
%     \begin{tabular}{l l l l l} 
%     \hline 
%     \emph{Framework} & \emph{Architecture} &\emph{Accuracy in \%} &Accuracy in \% & \emph{Standard deviation} \\
%      & &\emph{Average} & \emph{Maximum}& \\    
%      \hline 
%      \emph{ODESA Hardware (this work)} & ODESA 4\_\_6\_3\_\_3 & 79.5 &86.6 & 0.072\\ 
%      \hline 
%      \emph{ODESA Software \cite{YESH}} & ODESA 4\_\_6\_3\_\_3 & 82.8 &87.0 & 0.017\\ 
%      \hline 
%      \emph{Fully Connected Neural Network} & 4-input $\times$ 3-output  neurons & 81.7 &97.7 & 0.140 \\ 
%      \hline 
%     \end{tabular}
%   \end{center}
%   }
% \end{table*}
\begin{table*}[ht]\small{
  \begin{center}
    \caption{Iris dataset Accuracy test comparison }
    \label{tab:iris_compare}
    \begin{tabular}{l l l} 
    \hline 
    
    \emph{Framework} & \emph{Architecture} &\emph{Accuracy in \%} \\
     \hline
     \emph{ODESA Hardware model (this work)} & ODESA 4\_\_6\_3\_\_3 & 0.795 $\pm$ 0.072 \\ 
     \hline 
     \emph{ODESA Software model (this work)} & ODESA  4\_\_6\_3\_\_3 & 0.828 $\pm$ 0.017 \\ 
     \hline 
     \emph{ODESA Software model \cite{YESH}} & ODESA 20\_\_10\_3\_\_3 & 0.956 $\pm$ 0.010 \\ 
     \hline 
    \end{tabular}
  \end{center}
  }
\end{table*}
% The software version of ANN achieves the highest accuracy among the three given that it has access to full precision weights, activation values, and gradients with respect to the loss. Unlike the batch gradient descent method used in the ANN, ODESA performs online learning using only local variables and without access to the exact gradients to move the weights in the direction of minimizing a loss function. The learning in ODESA is similar to clustering methods and the limitations of ODESA networks also include the inability to learn negative weights unlike ANNs trained by back-propagation \cite{YESH}. 
The hardware version of ODESA does not show a considerable drop in accuracy compared to the software version even with the usage of non-normalized integer-based weights and fixed-step weight and threshold updates as described in Equations \ref{eq:wupdate}, \ref{eq:wupdate_l2}, and \ref{eq:adaptive_d}. But it does have a larger standard deviation compared to the software version, which is a result of using  a fixed-length integer numbering system and fixed-step updates that can limit the convergence.
%Our test results are detailed in Table \ref{tab:iris_compare}. We used the same dataset on the ODESA software version and a fully connected Neural Network to compare the ODESA hardware training performance.The software version of ODESA is the original algorithm from \cite{YESH} that used floating-point operations and normalized weights and input time surface for calculating the dot products of the neurons. The dot product in the software version is always bounded between 0 and 1.   
%The ODESA software version configuration is similar to the hardware, i.e., ODESA 4\_\_6\_3\_\_3. The fully connected Neural Network is simulated using a Keras back-end, with four neurons at the input level with a ReLu activation function and three neurons at the output level with a softmax activation function, a categorical\_crossentropy loss function, and an Adam optimizer.
The results show that our hardware ODESA and training algorithms perform very closely to the software version of ODESA.

\section{Conclusion}\label{sec:Conclusion}
For the first time, we presented an FPGA implementation of ODESA SNN that can be trained online in a supervised manner on hardware reset. The training data is stored in the internal RAM of the FPGA device and will be used on hardware restart to assign SNN parameters. This trainable hardware is efficient in terms of hardware resources and computing costs, making it appealing for streaming pattern detection applications, e.g., intrusion detection on IoT devices. The architecture can asynchronously update the neuron parameters at each layer independent of other layers in a network. All the communication in the hardware is event-based and via binary spikes. The architecture is capable of performing on-chip online learning and is a promising next step toward building energy-efficient continual learning edge devices.

Our work aims to draw attention to designing autonomous hardware that makes decisions based on receiving sensory inputs. Our approach could be extended to handle more complex pattern detection and classification tasks in near real-time. 
%%%%%%
\section*{Acknowledgment}\label{sec:ack}
This research is supported by the Commonwealth of Australia as represented
by the Defense Science and Technology Group of the Department of Defense.
%%%%%%
\ifCLASSOPTIONcaptionsoff
  \newpage
\fi
\bibliographystyle{IEEEtran}
\bibliography{IEEEabrv,Bibliography}

%\if\switchoff
% ==== SWITCH OFF the BIO for submission
\begin{IEEEbiography}[{\includegraphics[width=1in,height=1.25in,clip,keepaspectratio]{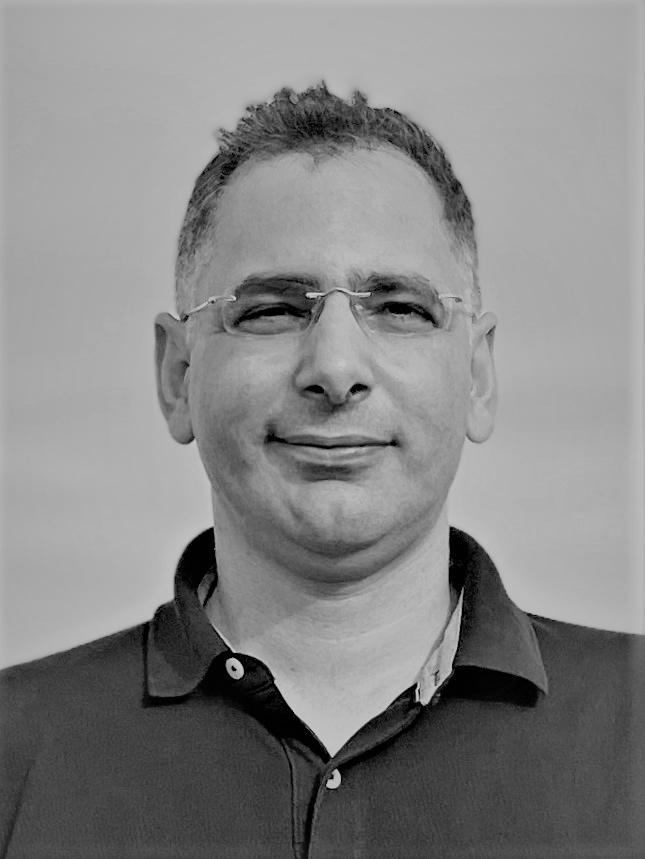}}]{ALI MEHRABI}
received his M.Sc. degree in Electronics Engineering from Shahid Beheshti University and Ph.D. degree in Computer Science from Macquarie University Sydney, Australia. Currently, he is a postdoctoral research fellow at International Center for Neuromorphic Systems, Western Sydney University.
His research interests include high performance computing hardware systems, event-based and neuromorphic signal processing. 
\end{IEEEbiography}
\begin{IEEEbiography}[{\includegraphics[width=1in,height=1.25in,clip,keepaspectratio]{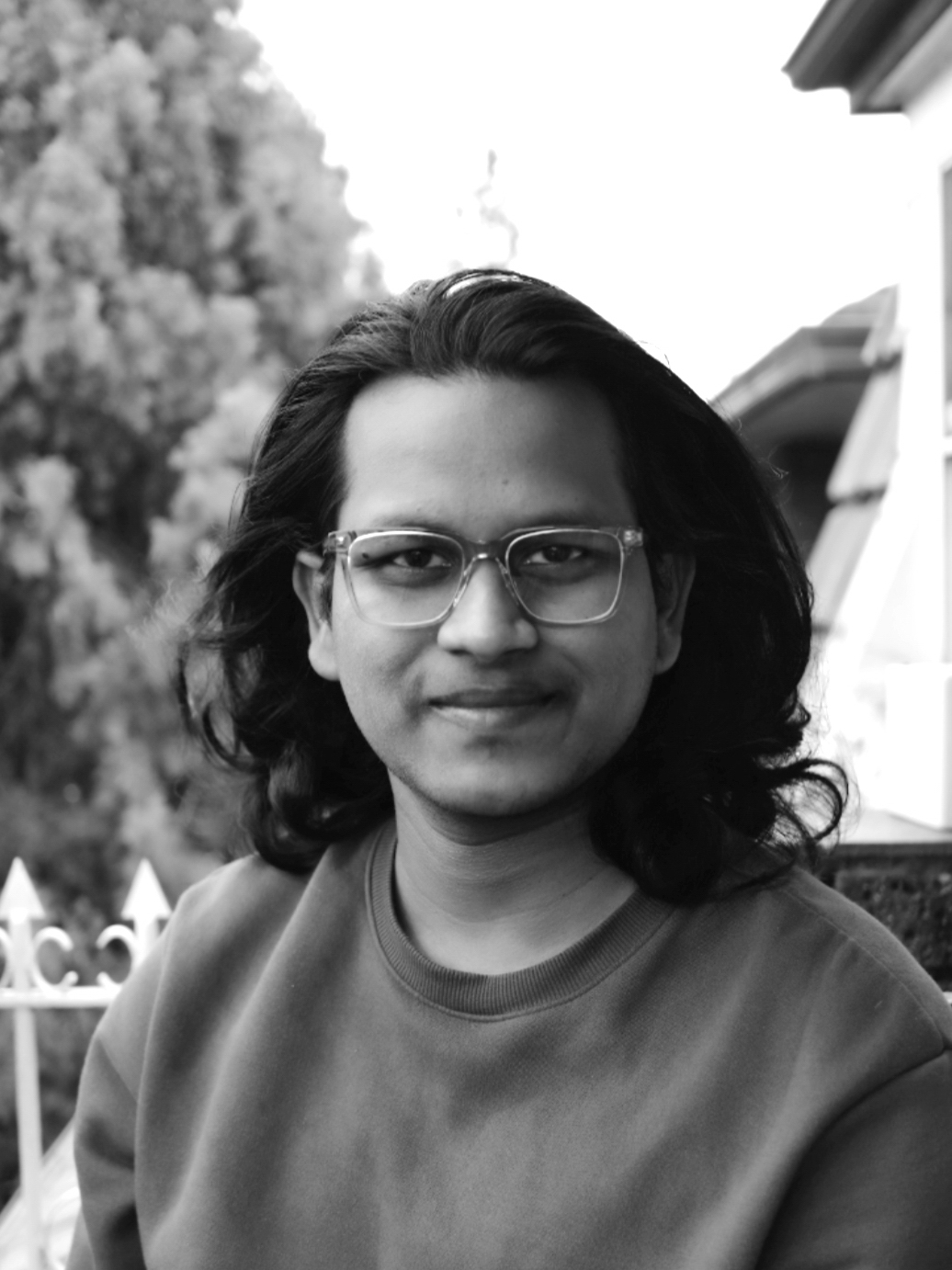}}]{YESHWANTH BETHI}
received the B.Tech. degree in electrical engineering from the Indian
Institute of Technology Bombay (IIT B), India, in 2016. He is currently pursuing the Ph.D. degree in event-based neural architectures with the International Centre for Neuromorphic Systems, Western Sydney University, Sydney, NSW, Australia. He was a Research Assistant with the Neuronics Laboratory, Indian Institute of Science (IISc), Bengaluru, India, from 2017 to 2019.
\end{IEEEbiography}
\begin{IEEEbiography}[{\includegraphics[width=1in,height=1.25in,clip,keepaspectratio]{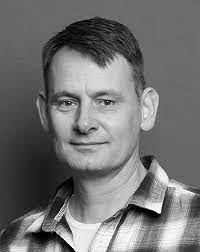}}]{ANDR\'E VAN SCHAIK}
received the M.Sc. degree in electrical engineering from the University of Twente, Enschede, The Netherlands, in 1990, and the Ph.D. degree in neuromorphic engineering from the Swiss Federal Institute of Technology (EPFL), Lausanne, Switzerland, in 1998. From 1991 to 1994, he was a Researcher with the Swiss Centre for Electronics and Microtechnology (CSEM), where he has been developing the first commercial neuromorphic chip—the optical motion detector used in Logitech trackballs, since 1994.
In 1998, he was a Postdoctoral Research Fellow with the Department of Physiology, University of Sydney, and in 1999, he became a Senior Lecturer
with the School of Electrical and Information Engineering and a Reader, in 2004. In 2011, he became a Full Professor at Western Sydney University, where he is currently Director of the International Center for Neuromorphic Systems. He has authored more than 250 articles and is an inventor of more than 35 patents.
He has founded three technology start-ups. His research interests include neuromorphic engineering, encompassing neurophysiology, computational neuroscience, software and algorithm development, and electronic hardware design. He is a pioneer of Neuromorphic Engineering and a Fellow of the IEEE for contributions to Neuromorphic Circuits and Systems.
\end{IEEEbiography}
\begin{IEEEbiography}[{\includegraphics[width=1in,height=1.25in,clip,keepaspectratio]{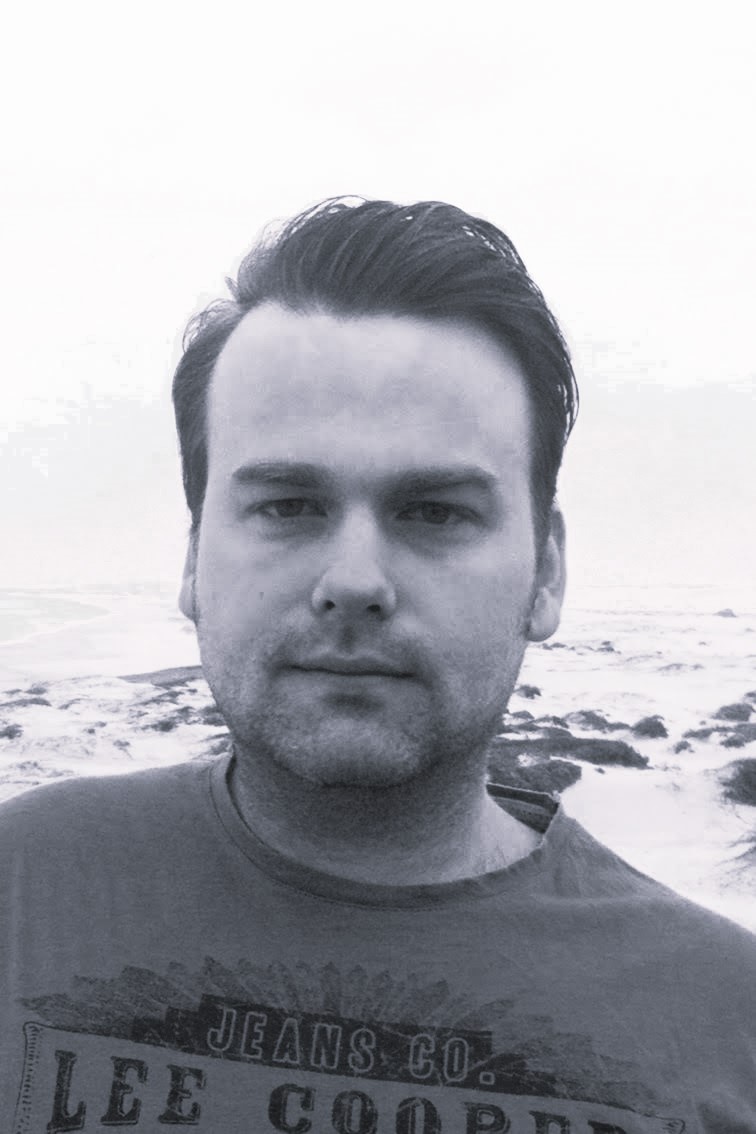}}]{ANDREW WABNITZ}
received the B.E. (Hons) degree in electrical engineering from the University of Adelaide, Australia, in 2005, and the Ph.D. degree from the University of Sydney, Australia, in 2013. Andrew has over 10 years experience across industry, academia and defence working on projects involving embedded system design software development and FPGA design for applications in biomedical devices, cyber security and space. He is currently a senior researcher of Cognitive Technologies in the Defence Science and Technology Group, Department of Defence, Australia. His primary research interests include neuromorphic computing, bio-inspired algorithm design, and the application of this technology to real-world problems.
\end{IEEEbiography}
\begin{IEEEbiography}[{\includegraphics[width=1in,height=1.25in,clip,keepaspectratio]{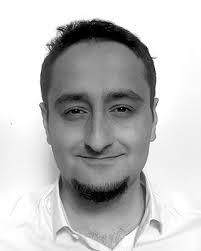}}]{SAEED AFSHAR}
received the B.Eng. degree in electrical engineering from the
University of New South Wales, Sydney, NSW, Australia, and the M.Eng. degree in electrical engineering from the Western Sydney University, Sydney. He is currently a Lecturer with the International Center for Neuromorphic Systems, Western Sydney University. His research interests include event-based vision and audio processing for neuromorphic hardware.
\end{IEEEbiography}
%\fi
\vfill
\end{document}